\pdfoutput=1
\documentclass{article}

\usepackage{iclr2027_conference,times}

\usepackage[T1]{fontenc}
\usepackage[utf8]{inputenc}
\usepackage{microtype}

\usepackage{amsmath}
\usepackage{amssymb}
\usepackage{mathtools}

\usepackage{algorithm}
\usepackage{algpseudocode}
\usepackage{xcolor}
\usepackage{tcolorbox}
\tcbuselibrary{skins,breakable}
\usepackage{needspace}

\usepackage{graphicx}
\usepackage{capt-of}
\usepackage{booktabs}
\usepackage{longtable}
\usepackage{multirow}
\usepackage{siunitx}
\usepackage{colortbl}
\definecolor{lptrow}{RGB}{246,241,247}
\definecolor{groupline}{RGB}{190,190,190}
\colorlet{takeawaybg}{lptrow}
\definecolor{takeawayaccent}{HTML}{7B3294}

\newcounter{takeaway}
\newtcolorbox{takeawaybox}{%
  enhanced,
  breakable,
  parbox=false,
  colback=takeawaybg,
  colframe=takeawayaccent,
  coltext=black,
  fontupper=\normalsize,
  frame hidden,
  boxrule=0pt,
  borderline west={2pt}{0pt}{takeawayaccent},
  boxsep=0pt,
  left=7pt,
  right=7pt,
  top=4pt,
  bottom=4pt,
  arc=1pt,
  before skip=7pt,
  after skip=10pt,
}
\newcommand{\newtakeaway}[1]{%
  \refstepcounter{takeaway}%
  \begin{takeawaybox}%
    \textcolor{takeawayaccent}{\textbf{Takeaway \thetakeaway.}}\enspace #1%
  \end{takeawaybox}%
}

\usepackage{stfloats}\fnbelowfloat   %

\usepackage[hidelinks]{hyperref}

\usepackage{url}

\newcommand{\emb}{\operatorname{emb}}
\newcommand{\lmhead}{\operatorname{lm\_head}}

\newcommand{\Std}{\textsc{std}}                     %
\newcommand{\DeepStd}{\textsc{deepstd}}             %
\newcommand{\LT}{\textsc{lt}}                       %
\newcommand{\PLT}{\textsc{plt}}                     %
\newcommand{\LPT}{\textsc{lpt}}                      %
\newcommand{\LTtwo}{\textsc{lt2}}                    %
\newcommand{\HLPT}{\textsc{h-lpt}}                   %
\newcommand{\SPS}{\textsc{sps}}                     %
\newcommand{\Loopie}{\textsc{loopie}}                %
\newcommand{\MemorySharing}{Memory Sharing}        %
\newcommand{\stack}{\mathcal{M}}                    %
\newcommand{\numrec}{R}                              %
\newcommand{\wpred}{\omega}                         %
\newcommand{\Kcache}{\mathcal{C}}                    %
\newcommand{\attset}{\mathcal{A}}                    %

\newcommand{\vx}{\mathbf{x}}

\iclrfinalcopy  %
\title{The Surprising Effectiveness\\of Shared Memory in Looped Transformers}
\author{%
  Giovanni Monea$^{1*}$ \quad Keshav Ramji$^{2}$ \quad Yousef El-Kurdi$^{2}$ \quad Luis A. Lastras$^{2}$ \\
  \textbf{Yoav Artzi$^{1}$ \quad Nathan Godey$^{1\dagger}$ \quad Ram\'on Fernandez Astudillo$^{2\dagger}$} \\
  $^{1}$Cornell University \quad $^{2}$IBM Research
}

\begin{document}
\setlength{\textfloatsep}{10pt plus 2pt minus 4pt}   %
\maketitle
\lhead{Preprint}  %
{\renewcommand{\thefootnote}{\fnsymbol{footnote}}%
 \footnotetext[1]{Work done during an internship at IBM Research. Correspondence to: \texttt{giovanni@cs.cornell.edu}}%
 \footnotetext[2]{Equal advising}}

\begin{abstract}
Looped Transformers apply the same layers several times per token, adding compute to improve quality without more parameters. Each recursion, however, writes its own key-value cache, so memory still grows with compute. Inference-time techniques can shrink this cache at a cost in quality. We pretrain looped language models to share memory: only the first recursion writes a cache, and later recursions read it while keeping a short window of their own. Surprisingly, we find that sharing memory does not cost quality and instead improves it. At 150M--1B parameters, our Looped Prediction Transformer (\LPT{}) and its hybrid variant set a new quality--memory frontier for looped models: with five recursions, the hybrid lowers validation perplexity on FineWeb-Edu by $1.12$--$1.82$ relative to a same-size standard Transformer while using $76$--$79\%$ less context memory. Through an extensive analysis, we investigate why memory sharing helps. Shared and local memory develop different representations, and later recursions attend mostly to the shared memory, which also acts as a gradient highway to the first recursion.
\end{abstract}

\definecolor{figcardbg}{HTML}{E2D5E8}   %
\definecolor{figcardrule}{HTML}{B9ACC2} %
\begin{figure}[H]
  \centering
  \setlength{\abovecaptionskip}{4pt}
  \begin{tcolorbox}[enhanced, frame hidden, boxrule=0pt, arc=6pt, colback=figcardbg, boxsep=0pt,
                    left=7pt, right=7pt, top=4pt, bottom=3pt, width=\linewidth,
                    sidebyside, sidebyside align=center, lefthand width=0.41\linewidth, sidebyside gap=8pt,
                    segmentation style={color=figcardrule, line width=0.6pt, solid}]
    \centering
    \includegraphics[width=\linewidth]{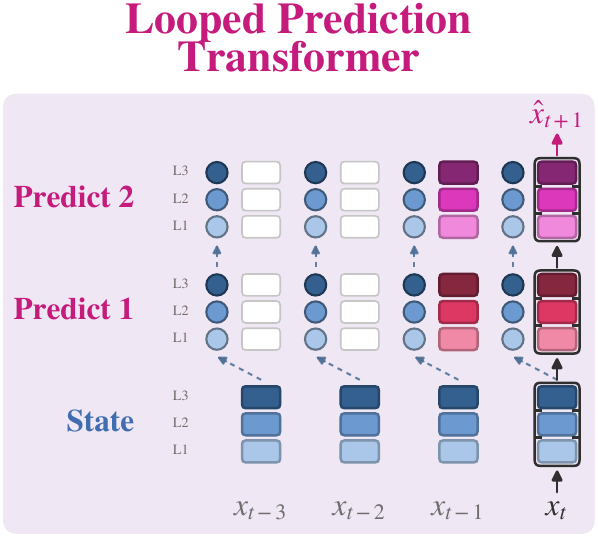}\\[1pt]
    \includegraphics[width=0.98\linewidth]{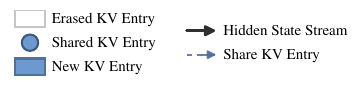}
    \tcblower
    \centering
    \includegraphics[width=\linewidth]{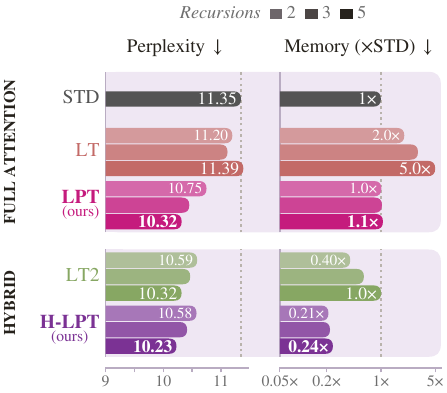}
  \end{tcolorbox}
  \caption{\textbf{Shared memory not only saves memory, it also improves quality.} \textbf{Left:} in our Looped Prediction Transformer (\LPT{}), later recursions read the KV cache written by the first recursion and keep only a short window of their own. \textbf{Right:} perplexity and context memory of $1$B-parameter models with two to five recursions, against a parameter-matched standard Transformer (\Std{}). Each of our models adds shared memory to the looped baseline above it: \LPT{} to the looped Transformer \LT{}, and \HLPT{} to the gated DeltaNet hybrid \LTtwo{}.}
  \label{fig:overview}
  \label{fig:architecture}
  \label{fig:quality-memory-by-size}
\end{figure}

\section{Introduction}

Looped Transformers (\LT{}) repeatedly apply the same layers several times before predicting the next token, adding computation without adding parameters~\citep{dehghani2019universal,ouro2025}. In principle, they trade time for memory, matching the compute of a larger model with far fewer parameters. In standard formulations, however, context memory still grows with recursion depth, as each iteration keeps a separate key-value (KV) cache~\citep{geiping2025huginn,ouro2025}. At long contexts, this context memory can grow larger than the weights memory itself, undoing the very reason to use a looped model.

Sharing caches across recursions during decoding can save substantial memory with little quality loss~\citep{geiping2025huginn,ouro2025}. So far, this reuse has been studied mainly as a way to save memory. Yet it also shows that memory written by one recursion can serve the others, even in models never trained to share it. This raises a question: \emph{if sharing can work untrained, what can it do when models are trained for it?} To answer it, we train looped models with memory sharing from the start, investigating it as an architectural inductive bias.

We study a simple recipe for \emph{\MemorySharing{}} in looped Transformers (\autoref{fig:overview}, left). At each full-attention layer, later recursions attend to both their own KV entries and the cache written by that layer in the first recursion, within the same attention operation. To further save memory, we truncate the cache of every recursion after the first to a short sliding window. The resulting Looped Prediction Transformer (\LPT{}) keeps \LT{}'s sequential recurrence and uses context memory close to a standard (non-looped) Transformer's. Applied to the full-attention layers of a hybrid looped Transformer~\citep{deng2026lt2}, the same rule yields the Hybrid Looped Prediction Transformer (\HLPT{}). We pretrain all models from scratch at $150$M to $1$B parameters. Because later recursions give up most of their own memory, one would expect \MemorySharing{} to cost some quality, as untrained sharing does.

Surprisingly, \MemorySharing{} \emph{improves} quality over full-cache \LT{} even when each later recursion's own cache covers less than $2\%$ of the context. Its advantage grows with depth: \LT{} stops improving in perplexity after three recursions, while \LPT{} improves through five. Across sizes, \LPT{} lowers validation perplexity by up to $1.09$ nats and raises downstream accuracy by up to $4.8$ percentage points over a matched-depth \LT{}, with up to $4.7\times$ less context memory and nearly the same inference FLOPs. Sharing also improves the hybrid baseline: at five recursions, \HLPT{} outperforms a same-size standard Transformer in both language modeling and downstream accuracy while using $76$--$79\%$ less context memory, moving the quality--memory frontier for looped models (\autoref{fig:accuracy-memory-by-size}).

These results raise a further question: \emph{why does memory sharing improve quality?} We find that later recursions rely heavily on the shared memory. \LPT{} outperforms \LT{} at every window size, even when later recursions keep no cache of their own, and later recursions attend mostly to shared memory even when their own cache covers the full context. Shared and local memory also develop distinct key representations, suggesting that they serve different purposes. Reading shared memory sends gradients directly to the first recursion, helping it learn what to store. This gradient highway bypasses the intermediate recursions: in \LT{}, the magnitude of the gradient reaching the first recursion falls sharply with depth, while \LPT{} recovers much of it through the shared memory. When we block this path, the model still reads shared memory but no longer matches \LPT{}'s quality and stability.

\section{Looped Transformers}

A standard \LT{} shares its weights across recursions, while each recursion computes its own hidden states and KV entries. We use Ouro's full-stack formulation~\citep{ouro2025}, which builds on earlier recurrent-depth architectures~\citep{dehghani2019universal,geiping2025huginn}.

Let $\stack^{L}=\mathcal{T}_{\theta}^{(L)}\circ\cdots\circ\mathcal{T}_{\theta}^{(1)}$ denote the shared stack of $L$ Transformer blocks. For a sequence of length $T$ and hidden dimension $d$, let $h_{\mathrm{in}}^{(t)}\in\mathbb{R}^{T\times d}$ and $h^{(t)}$ be the residual streams entering and leaving recursion $t$. The stack is applied $\numrec$ times:
\begin{equation}
  \begin{aligned}
    h_{\mathrm{in}}^{(1)} &= \emb(\vx), \\
    h^{(t)} &= \stack^{L}\!\big(h_{\mathrm{in}}^{(t)}\big), \\
    h_{\mathrm{in}}^{(t+1)} &= \mathcal{N}_{\mathrm{rec}}\!\big(h^{(t)}\big), \\
    \text{logits}^{(t)} &= \lmhead\!\big(\mathcal{N}_{\mathrm{out}}(h^{(t)})\big),
      \qquad t=1,\dots,\numrec .
  \end{aligned}
  \label{eq:lt-recurrence}
\end{equation}
Here $\mathcal{N}_{\mathrm{rec}}$ and $\mathcal{N}_{\mathrm{out}}$ are separately parameterized RMSNorms. Both act on $h^{(t)}$: the recurrent norm prepares the next recursion's input, while the readout norm feeds the language-modeling head. The effective depth is $L\cdot\numrec$, with the parameters of a single $L$-layer stack.

Each layer computes its keys and values from the hidden states it receives at the current recursion. For a fixed layer, let $\Kcache^{(t)}$ denote the cache written by recursion $t$, with entry $(k_j^{(t)},v_j^{(t)})$ at position $j$. We suppress the layer index throughout: for vanilla LT, each layer reads only its own entries. A query at position $i$ reads the causal prefix of its recursion's cache, giving the attention set
\begin{equation}
  \attset_{\text{\LT}}^{(t)}(i)
    = \Kcache_{\le i}^{(t)}
    = \{(k_j^{(t)},v_j^{(t)}):j\le i\}.
  \label{eq:lt-cache}
\end{equation}
At inference, each layer therefore retains $\numrec$ full-context caches, one per recursion, and context memory grows linearly with recursion depth even though the parameter count stays fixed. Some of these caches can be dropped or merged during decoding~\citep{geiping2025huginn,ouro2025}, but the model is still trained with a separate cache for every recursion.

Hybrid architectures reduce this memory cost by replacing most full-attention layers with linear attention, whose recurrent state does not grow with context length~\citep{katharopoulos2020lineartransformers,schlag2021fastweights,yang2025gateddeltanet}. \LTtwo{}~\citep{deng2026lt2} loops such a stack, using Gated DeltaNet layers and one full-attention layer per five. This reduces the number of full-attention caches, but each remaining layer still needs a separate full-context cache at every recursion.

\section{Looping with Shared Memory}

We study \emph{\MemorySharing{}}, in which later recursions read the memory written by the first recursion, as an inductive bias for looped models. The first recursion runs as in \LT{} and writes a full-context cache $\Kcache^{(1)}$ at each layer, and every later recursion attends to this cache together with its own KV entries, using a single softmax.

To also keep looping memory-efficient, we restrict each later recursion's own cache to a short window. The first recursion therefore acts as a memory phase that builds the model's only full-context memory, while later recursions use it to refine the prediction. We refer to the first recursion as the \emph{state recursion} and to the later ones as \emph{prediction recursions}, and call the resulting model the Looped Prediction Transformer (\LPT{}; \autoref{fig:architecture}). Like \LT{}, \LPT{} applies the same $L$-layer stack $\numrec$ times and passes the final output to the language-modeling head (\autoref{eq:lt-recurrence}).

Let $\wpred$ be the number of preceding positions kept in each prediction recursion's window, and write $\Kcache_{[i-\wpred,i]}^{(t)}$ for the entries from positions $i-\wpred$ through $i$, including the current entry. The state recursion reads its own cache, as in \LT{} (\autoref{eq:lt-cache}). A prediction recursion reads the shared memory together with its own window. The entries read by a query at position $i$ in recursion $t$ are
\begin{equation}
  \attset_{\text{\LPT}}^{(t)}(i)=
  \begin{cases}
    \Kcache_{\le i}^{(1)},
      & t=1, \\
    \Kcache_{\le i}^{(1)} \cup \Kcache_{[i-\wpred,i]}^{(t)},
      & t=2,\ldots,\numrec .
  \end{cases}
  \label{eq:lpt-attention}
\end{equation}
In prediction recursions, the two sets are attended to using a single softmax.

Setting $\wpred=T$ gives every prediction recursion a full-context cache of its own, as in \LT{}, and leaves access to shared memory as the only difference between the two models; we use this setting in \autoref{sec:analysis} to separate the effect of sharing from that of the short window.

\MemorySharing{} also applies to looped hybrid Transformers, which interleave linear-attention and full-attention layers~\citep{yang2025gateddeltanet,deng2026lt2}. The first recursion runs the hybrid as usual. In later recursions, each linear-attention layer keeps its own recurrent state, which is small and fixed in size, while each full-attention layer applies \autoref{eq:lpt-attention}, as in \LPT{}. Applied to \LTtwo{}~\citep{deng2026lt2}, this gives \HLPT{}.

\paragraph{Efficiency and Costs} In training, we implement \autoref{eq:lpt-attention} as a FlexAttention~\citep{dong2024flexattentionprogrammingmodel} block mask over the concatenated shared and local entries, which skips the fully masked blocks, so each query attends over the shared prefix and its $\wpred$ window entries only (\autoref{app:experimental-details}).
  
During decoding, each prediction recursion keeps a sliding buffer of at most $\wpred$ preceding entries per layer; its current entry participates in attention before entering the buffer. In cached token positions across $L$ full-attention layers, the inference-time KV cache sizes are
\begin{equation}
  M_{\mathrm{KV}}^{\text{\LT}} = L\,\numrec\,T,
  \qquad
  M_{\mathrm{KV}}^{\text{\LPT}} = L\left[T+(\numrec-1)\wpred\right],
  \label{eq:lpt-memory}
\end{equation}
where the common head dimensions and datatype size are omitted. \LPT{} uses the same parameters and recursion count as a matched-depth \LT{}, but performs slightly more attention because each prediction recursion also reads its local window. At a full context of $T=4{,}096$ with $\wpred=64$ and five recursions, these windows increase the number of attended KV entries per generated token by only $(\numrec-1)\wpred/(\numrec T)=1.25\%$, while reducing the full-attention cache relative to \LT{} by $4.7\times$.

\begin{table}[!t]
    \centering
    \setlength{\abovecaptionskip}{0pt}\setlength{\belowcaptionskip}{6pt}
  \caption{\textbf{\LPT{} and \HLPT{} reach lower perplexity than their looped baselines with much smaller caches.} Main results across S, M, and L: validation perplexity, mean five-task accuracy (\%), and context memory and FLOPs relative to same-size \Std{}. In the \Std{} rows, $\numrec=1$ is the standard Transformer, and $\numrec>1$ is \DeepStd{}. Elsewhere, $\numrec$ counts model passes. \textbf{Bold} and \underline{underline} mark the best and second-best within each size and column group; our rows are shaded. We replicate the S results with three additional seeds and find the gains of \MemorySharing{} statistically significant ($p<0.01$, \autoref{app:seeds}).}
  \label{tab:main-results}
    \scriptsize
    \sisetup{mode=text, detect-weight=true}
    \setlength{\tabcolsep}{2pt}
    \renewcommand{\arraystretch}{1.15}
    \resizebox{\textwidth}{!}{%
    \begin{tabular}{@{}c l S[table-format=2.2] S[table-format=2.1] S[table-format=2.2] S[table-format=2.1] S[table-format=1.2] S[table-format=1.2] S[table-format=2.2] S[table-format=2.1] S[table-format=1.2] S[table-format=1.2] S[table-format=2.2] S[table-format=2.1] S[table-format=1.2] S[table-format=1.2]@{}}
        \toprule
        & & \multicolumn{2}{c}{$\numrec=1$} & \multicolumn{4}{c}{$\numrec=2$} & \multicolumn{4}{c}{$\numrec=3$} & \multicolumn{4}{c}{$\numrec=5$} \\
        \cmidrule(lr){3-4} \cmidrule(lr){5-8} \cmidrule(lr){9-12} \cmidrule(lr){13-16}
        Size & Model & {PPL $\downarrow$} & {Acc. $\uparrow$} & {PPL $\downarrow$} & {Acc. $\uparrow$} & {KV $\downarrow$} & {FLOPs $\downarrow$} & {PPL $\downarrow$} & {Acc. $\uparrow$} & {KV $\downarrow$} & {FLOPs $\downarrow$} & {PPL $\downarrow$} & {Acc. $\uparrow$} & {KV $\downarrow$} & {FLOPs $\downarrow$} \\
        \midrule
         & \Std{} & 16.83 & 51.5 & 17.07 & 50.2 & 1.50 & {\bfseries 1.19} & 17.27 & 50.7 & 1.88 & {\bfseries 1.31} & 17.69 & 50.4 & 2.50 & {\bfseries 1.54} \\
         & \Loopie{} &  &  & 16.58 & 51.6 & 2.00 & 1.83 & 16.50 & 51.9 & 3.00 & 2.66 & 16.56 & 52.1 & 5.00 & 4.32 \\
         & \LT{} &  &  & 16.17 & 51.5 & 2.00 & 1.83 & 15.98 & 52.1 & 3.00 & 2.66 & 16.00 & 52.4 & 5.00 & 4.32 \\
         & \PLT{} &  &  & 15.84 & {\bfseries 52.7} & 1.02 & 1.84 & 15.55 & 52.9 & 1.03 & 2.67 & 15.49 & 53.5 & 1.06 & 4.34 \\
         & \LTtwo{} &  &  & {\underline{15.76}} & 51.9 & {\underline{0.35}} & {\underline{1.29}} & 15.47 & 52.5 & {\underline{0.52}} & {\underline{1.86}} & 15.16 & 53.7 & {\underline{0.87}} & {\underline{2.98}} \\
        \rowcolor{lptrow}
         & \textbf{\LPT{}} &  &  & {\bfseries 15.73} & 51.9 & 1.02 & 1.84 & {\bfseries 15.28} & {\bfseries 54.0} & 1.03 & 2.67 & {\bfseries 14.91} & {\bfseries 54.9} & 1.06 & 4.34 \\
        \rowcolor{lptrow}
        \multirow{-7}{*}{\textbf{S}} & \textbf{\HLPT{}} &  &  & {\bfseries 15.73} & {\underline{52.4}} & {\bfseries 0.18} & {\underline{1.29}} & {\underline{15.39}} & {\underline{53.1}} & {\bfseries 0.19} & {\underline{1.86}} & {\underline{15.01}} & {\underline{54.6}} & {\bfseries 0.21} & {\underline{2.98}} \\
        \addlinespace[3pt]
        \arrayrulecolor{groupline}\cmidrule[0.3pt](l{2pt}r{2pt}){1-16}\arrayrulecolor{black}
        \addlinespace[3pt]
         & \Std{}  & 13.79 & 56.1 & 14.24 & 55.5 & 1.44 & {\bfseries 1.13} & 14.47 & 55.4 & 1.78 & {\bfseries 1.24} & 14.88 & 54.4 & 2.34 & {\bfseries 1.43} \\
         & \Loopie{} &  &  & 13.69 & 56.9 & 2.00 & 1.92 & 13.75 & 55.8 & 3.00 & 2.84 & 13.74 & 56.6 & 5.00 & 4.69 \\
         & \LT{} &  &  & 13.28 & 57.5 & 2.00 & 1.92 & 13.13 & 59.4 & 3.00 & 2.84 & 13.34 & 57.9 & 5.00 & 4.69 \\
         & \PLT{} &  &  & 13.00 & {\underline{59.1}} & 1.02 & 1.93 & 12.68 & 60.0 & 1.03 & 2.85 & 12.65 & 60.5 & 1.06 & 4.71 \\
         & \LTtwo{} &  &  & {\underline{12.81}} & {\underline{59.1}} & {\underline{0.35}} & {\underline{1.43}} & 12.63 & 59.5 & {\underline{0.52}} & {\underline{2.10}} & 12.43 & 60.2 & {\underline{0.87}} & {\underline{3.45}} \\
        \rowcolor{lptrow}
         & \textbf{\LPT{}} &  &  & 12.91 & {\bfseries 59.2} & 1.02 & 1.93 & {\underline{12.55}} & {\underline{60.2}} & 1.03 & 2.85 & {\underline{12.33}} & {\bfseries 61.6} & 1.06 & 4.70 \\
        \rowcolor{lptrow}
        \multirow{-7}{*}{\textbf{M}} & \textbf{\HLPT{}} &  &  & {\bfseries 12.80} & 58.5 & {\bfseries 0.18} & {\underline{1.43}} & {\bfseries 12.54} & {\bfseries 60.3} & {\bfseries 0.19} & {\underline{2.10}} & {\bfseries 12.31} & {\underline{61.4}} & {\bfseries 0.21} & {\underline{3.45}} \\
        \addlinespace[3pt]
        \arrayrulecolor{groupline}\cmidrule[0.3pt](l{2pt}r{2pt}){1-16}\arrayrulecolor{black}
        \addlinespace[3pt]
         & \Std{}  & 11.35 & 62.5 & 11.62 & 62.5 & 1.45 & {\bfseries 1.14} & 11.96 & 61.3 & 1.73 & {\bfseries 1.17} & 12.27 & 61.4 & 2.25 & {\bfseries 1.32} \\
         & \Loopie{} &  &  & 11.26 & 63.5 & 2.00 & 1.95 & 11.28 & 63.9 & 3.00 & 2.91 & 12.26 & 61.0 & 5.00 & 4.81 \\
         & \LT{} &  &  & 11.20 & 64.2 & 2.00 & 1.95 & 11.11 & 63.8 & 3.00 & 2.91 & 11.39 & 63.4 & 5.00 & 4.81 \\
         & \PLT{} &  &  & 10.95 & 64.8 & 1.02 & 1.96 & 10.75 & 66.8 & 1.03 & 2.92 & 10.59 & 66.4 & 1.06 & 4.83 \\
         & \LTtwo{} &  &  & {\underline{10.59}} & {\bfseries 66.5} & {\underline{0.40}} & {\underline{1.53}} & 10.47 & 65.8 & {\underline{0.60}} & {\underline{2.27}} & {\underline{10.32}} & 67.3 & {\underline{1.01}} & {\underline{3.75}} \\
        \rowcolor{lptrow}
         & \textbf{\LPT{}} &  &  & 10.75 & {\underline{66.3}} & 1.02 & 1.96 & {\underline{10.45}} & {\bfseries 67.3} & 1.03 & 2.92 & {\underline{10.32}} & {\bfseries 68.2} & 1.06 & 4.83 \\
        \rowcolor{lptrow}
        \multirow{-7}{*}{\textbf{L}} & \textbf{\HLPT{}} &  &  & {\bfseries 10.58} & 65.6 & {\bfseries 0.21} & {\underline{1.53}} & {\bfseries 10.41} & {\underline{66.9}} & {\bfseries 0.22} & {\underline{2.27}} & {\bfseries 10.23} & {\underline{68.1}} & {\bfseries 0.24} & {\underline{3.75}} \\
        \bottomrule
    \end{tabular}%
    }
\end{table}

\section{Experimental Setup}
\label{sec:experimental-setup}

\paragraph{Baselines}
To study memory sharing, we compare \LPT{} with \LT{} and \HLPT{} with \LTtwo{}. We also compare with the Parallel Loop Transformer (\PLT{})~\citep{wu2025parallel}, which parallelizes recursions and uses gated dual attention over the first recursion's shared cache and each later recursion's local cache. Other baselines are a standard Transformer (\Std{}), \Loopie{}, which repeats each layer before advancing to the next~\citep{gao2026loopies}, and \DeepStd{}, a deeper, narrower Transformer without weight sharing that matches the looped models in effective depth and approximately in parameters.

\paragraph{Architecture}
All models use a decoder-only Transformer backbone with RMSNorm, SwiGLU, rotary positional embeddings, and tied input and output embeddings. We train S, M, and L models with approximately $150$M, $450$M, and $1$B parameters; the main comparisons cover all three sizes, and most ablations and analyses use S and M. The context length is $4{,}096$. Unless otherwise specified, \LPT{} and \HLPT{} use a prediction window of $\wpred=64$. We study two, three, and five recursions. This range includes the two loop steps of \Loopie{}~\citep{gao2026loopies} and surrounds the four recursions of Ouro and \LTtwo{}~\citep{ouro2025,deng2026lt2}.\footnote{Nine recursions bring instability in several model families and limited accuracy gains, consistent with the loss spikes Ouro reports at eight~\citep{ouro2025}. We report these experiments in \autoref{app:nine-recursions}.} All models use nanoGPT's default optimizer settings~\citep{Karpathy2022}. Model dimensions, normalization, and training details are in \autoref{app:experimental-details}. 

\paragraph{Data and Training}
We pretrain on FineWeb-Edu~\citep{penedo2024fineweb} using $20$B tokens for S and M, and $50$B for L, roughly $2$--$7\times$ the Chinchilla compute-optimal token budgets~\citep{hoffmann2022chinchilla}. The main tables report one run per configuration, with the same initialization seed and data order across models. At S, we also train three more seeds per model and confirm that seed variation is far below the gaps we report (\autoref{app:seeds}).

We study two training objectives. \emph{Final loss} computes the next-token NLL at the final recursion only. \emph{Adaptive loss} follows Ouro~\citep{ouro2025}, weighting per-recursion NLLs with a learned per-token halting distribution and adding KL regularization. Unless otherwise specified, \LT{} uses adaptive loss and all other models use final loss.

\paragraph{Evaluation}
We report perplexity on $10$M held-out FineWeb-Edu tokens and mean zero-shot accuracy on ARC-Easy~\citep{clark2018arc}, HellaSwag~\citep{zellers2019hellaswag}, PIQA~\citep{bisk2020piqa}, SciQ~\citep{welbl2017sciq}, and LAMBADA~\citep{paperno2016lambada}, using the LM Evaluation Harness~\citep{eval-harness}. We compute context memory and decoding FLOPs per generated token analytically at a $4{,}096$-token context, relative to the same size \Std{}. Context memory includes the KV cache and GDN's recurrent state.

\begin{figure}[!t]
  \centering
  \includegraphics[width=\linewidth]{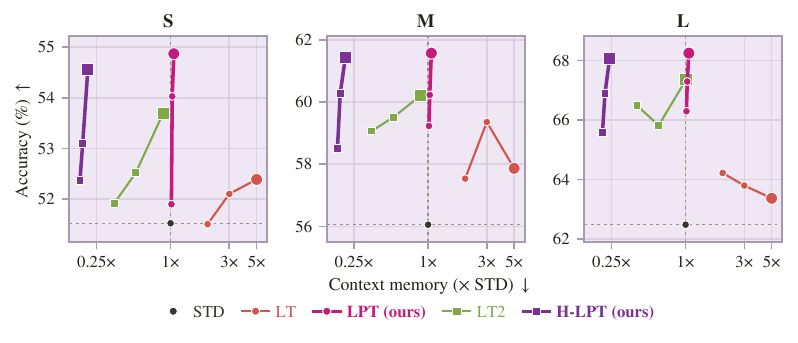}
  \caption{\textbf{\LPT{} and \HLPT{} reach higher accuracy with much smaller caches than the looped baselines.} Mean five-task accuracy vs. context memory at each size. Curves follow two, three, and five recursions, with larger markers at five. Dashed lines mark single-pass \Std{}.}
  \label{fig:accuracy-memory-by-size}
\end{figure}

\section{Results}
\label{sec:results}

\paragraph{\MemorySharing{} Improves Quality}
Restricting the memory written by later recursions could hurt prediction. Surprisingly, our \MemorySharing{} approach improves quality. At matched size and recursion count, \LPT{} lowers validation perplexity by $0.38$--$1.09$ and raises accuracy by $0.4$--$4.8$ percentage points over \LT{}, with $2.0$--$4.7\times$ less KV memory and under $1\%$ more FLOPs (\autoref{tab:main-results}). Measured on an H100, the memory savings are the same, so \LPT{} fits $2.0$--$4.8\times$ as many sequences in a batch as \LT{} (\autoref{app:wallclock}).
The perplexity gain grows with recursion count (\autoref{tab:main-results}): \LT{} stops improving beyond three recursions at every size, while \LPT{} continues to improve through five. At comparable cache sizes, \LPT{} also reaches lower perplexity than \PLT{}'s parallel formulation.  

When it comes to hybrid variants, \HLPT{} keeps this memory at $0.18$--$0.24\times$ \Std{}'s across recursion counts, while \LTtwo{}'s remaining full-attention caches still grow with depth, approaching \Std{}'s context memory at five recursions. Despite these $2$--$4\times $ memory gains, \HLPT{} matches or outperforms \LTtwo{} in terms of quality at every recursion count. At five recursions, it lowers perplexity by $0.09$--$0.15$ and raises accuracy by $0.8$--$1.2$ points over \LTtwo{}, with $4.1$--$4.2\times$ less context memory and about $0.1\%$ more inference FLOPs. 

Incidentally, we note that for low recursion counts, the hybrid \LTtwo{} improves on \LT{} while using less memory and compute. A factor that could explain this gap -- and the instability of \LT{} -- is Ouro's normalization, which we use for both \LT{} and \LPT{}. In \autoref{app:lt-normalization}, we explore other normalizations that make \LT{} more robust, but none of these variants matches \LPT{}.

\paragraph{More Quality from the Same Parameters}
Both memory-sharing models improve on \Std{} with little additional memory, or much less for \HLPT{} (\autoref{fig:accuracy-memory-by-size}). At five recursions, \LPT{} lowers perplexity by $1.03$--$1.92$ and raises accuracy by $3.4$--$5.7$ points with $6.25\%$ more context memory. \HLPT{} lowers perplexity by $1.12$--$1.82$ and raises accuracy by $3.1$--$5.6$ points with $76$--$79\%$ less memory. This smaller cache lets \HLPT{} fit $3.6$--$5.0\times$ as many sequences as \Std{} on one GPU. With three to five recursions, \HLPT{} lowers perplexity by $0.94$--$1.82$ while beating or staying close to \Std{}'s throughput ($1.4$--$1.5\times$ at three, $81$--$88\%$ at five, \autoref{app:wallclock}). Both models' accuracy gains grow from two to five recursions at every size. \LPT{} also outperforms \Loopie{} and \DeepStd{}, neither of which improves consistently with depth.

\newtakeaway{Through \MemorySharing{}, \LPT{} and especially \HLPT{} improves the quality--memory frontier for looped models (\autoref{fig:accuracy-memory-by-size}).}

\section{Why \MemorySharing{} Helps}
\label{sec:analysis}

We use S- and M-size models with two, three, and five recursions to study why this inductive bias improves quality (\autoref{app:analysis-details}).

\paragraph{\MemorySharing{} Stabilizes Deep Recursion}

\begin{figure}[!t]
  \centering
  \setlength{\abovecaptionskip}{4pt}\setlength{\belowcaptionskip}{6pt}
  \includegraphics[width=\linewidth]{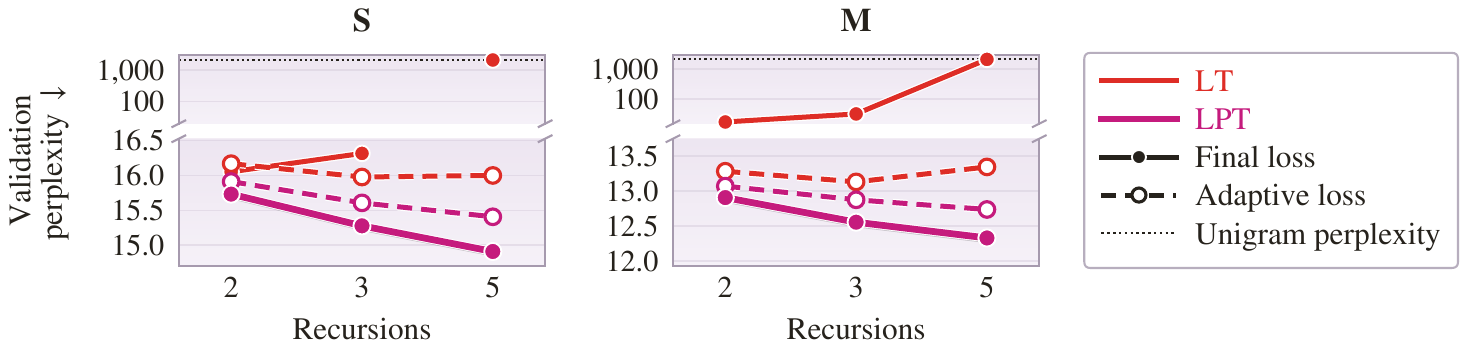}
  \caption{\textbf{\LPT{} trains best with final loss, while \LT{} consistently worsens with more recursions without adaptive loss.} Validation perplexity ($\downarrow$), collapsed runs in the upper strip on a log scale. Filled markers are final loss, hollow ones adaptive loss.}
  \label{fig:training-ablations}
  \par\vspace{\floatsep}
  \includegraphics[width=\linewidth]{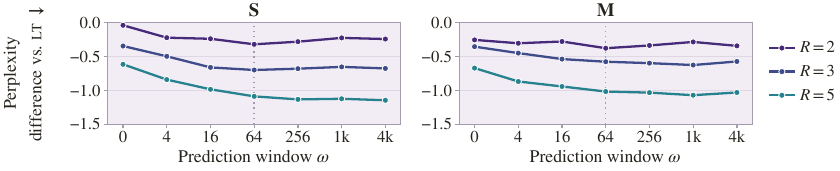}
  \caption{\textbf{A short prediction window is enough.} Validation perplexity of \LPT{} relative to \LT{} ($\downarrow$) vs. window size, \LT{} at its better objective per depth. The dotted line marks $\wpred=64$.}
  \label{fig:prediction-window}
\end{figure}

We compare final and adaptive loss in \LT{} and \LPT{} (\autoref{fig:training-ablations}). Without intermediate supervision, adding recursions consistently worsens \LT{}'s perplexity, and the model collapses to unigram-level loss at five recursions. \LPT{} stays stable, and final loss lowers its perplexity by $0.16$--$0.50$ compared with adaptive loss. Its intermediate recursions can refine the hidden state without having to predict early.

\newtakeaway{\LPT{} stays stable and trains best with final loss.}

\paragraph{Prediction Recursion Needs Local Memory}

We train \LPT{} with prediction windows ranging from $\wpred=0$ to the full context of $4{,}096$ tokens (\autoref{fig:prediction-window}). \LPT{} outperforms \LT{} at every window size, across both sizes and all tested depths. This includes $\wpred=0$, where prediction recursions keep no local history and read shared memory alongside only their current KV entry. At the other extreme, $\wpred=4{,}096$, \LT{} and \LPT{} differ only in access to shared memory. \LPT{} still lowers perplexity by $0.24$--$1.15$ relative to \LT{} with the better of its two training objectives at each depth, close to the $0.32$--$1.09$ gain at $\wpred=64$. Overall, window size has a modest effect on quality, and what really matters is the access to the shared memory. To balance quality and efficiency, we adopt $\wpred=64$ as the default for our memory sharing models.

\newtakeaway{Shared memory improves quality across prediction-window sizes, even when later recursions keep no local history.}

\paragraph{Later Recursions Attend More to Shared Memory}

\begin{figure}[!t]
  \centering
  \setlength{\abovecaptionskip}{5pt}
  \includegraphics[width=\linewidth]{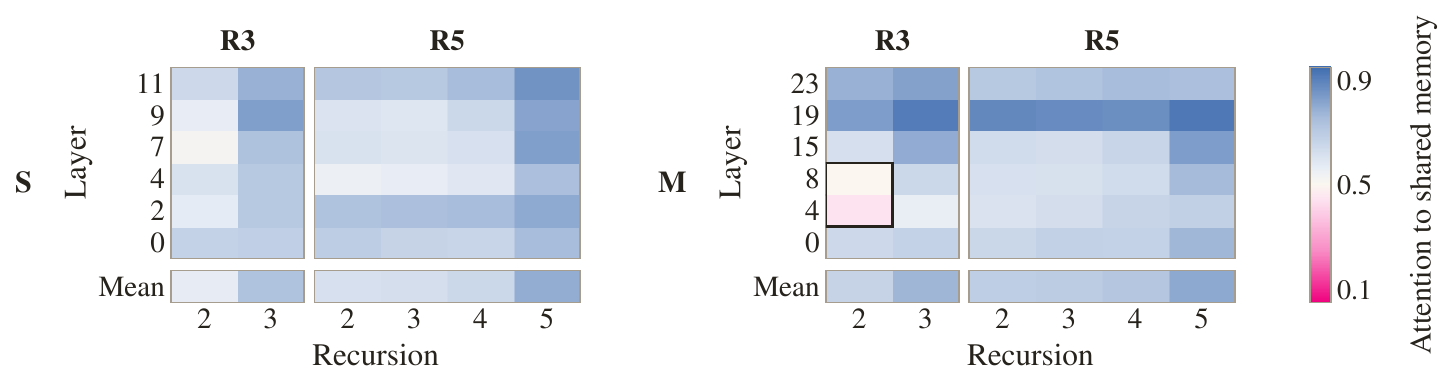}
  \caption{\textbf{Prediction recursions attend mostly to shared memory.} Attention mass assigned to shared KV entries in \LPT{}, averaged over queries and heads. For clarity, only six layers are displayed. The bottom row averages all layers.}
  \label{fig:state-attention-mass}
  \par\vspace{\floatsep}
  \includegraphics[width=\linewidth]{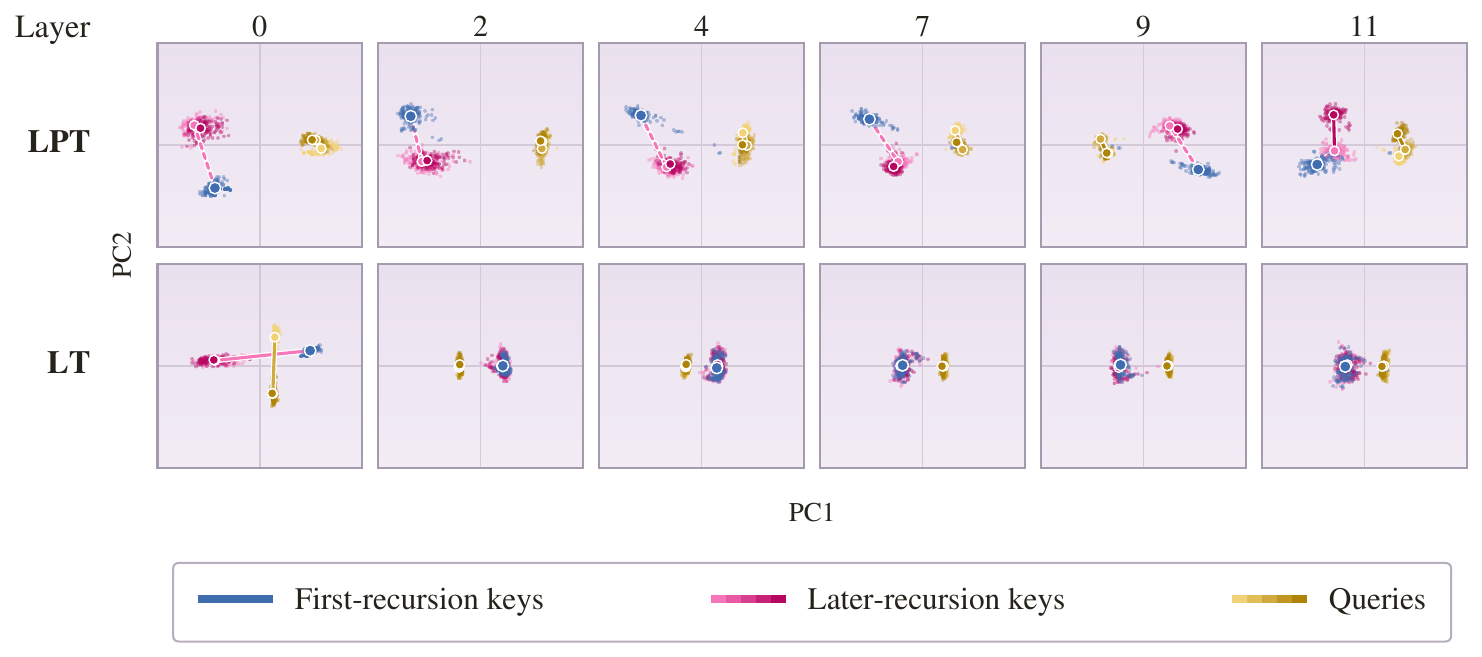}
  \caption{\textbf{Shared and local keys separate in \LPT{}.} Joint query/key PCA, fitted separately for each panel. Dots are tokens and large circles their means; darker shades mark later recursions.}
  \label{fig:qkv-pca}
\end{figure}

We next measure how much attention is assigned to the shared memory in later recursions. We evaluate each model on the same $4{,}096$-token validation sequence and average the attention assigned to shared keys over queries and heads (\autoref{fig:state-attention-mass} and \autoref{app:analysis-details}). Shared memory receives most attention at every prediction recursion in both sizes. Its share is $57$--$71\%$ in earlier prediction recursions and rises to $73$--$80\%$ in the final recursion.\footnote{Additional comparisons in the appendix show that shared memory still receives most attention with full prediction caches, while adaptive loss makes its use more uniform across recursions (\autoref{fig:state-attention-mass-conditions}).}

\newtakeaway{Later recursions attend more to shared memory than to their own caches, with a stronger preference when getting closer to the prediction.}

\paragraph{Shared and Local Memory Develop Different Representations}

Shared keys serve every later recursion, whereas local keys serve only the recursion that writes them. To see whether these roles are reflected in their representations, we compare S-size \LT{} and \LPT{} with three recursions, both trained with final loss, the only tested setting with at least three recursions where \LT{} trains with final loss without a substantial drop in quality (\autoref{fig:training-ablations}). We evaluate both models on the same $4{,}096$-token validation sequence and extract queries and keys before rotary encoding. We concatenate the attention heads and fit principal component analysis (PCA) jointly to queries and keys across tokens and recursions per model and layer (\autoref{app:analysis-details}).

Shared and local keys consistently form distinct clusters in \LPT{}, whereas keys from different recursions overlap more in \LT{} (\autoref{fig:qkv-pca}). In \autoref{app:analysis-details}, we show similar separation in \LPT{} across S and M at three and five recursions (\autoref{fig:lpt-pca-depth}). We also show that with full prediction caches, the clusters overlap more but still remain distinguishable, even though shared and local keys cover the same positions (\autoref{fig:qkv-pca-full}). Prior work attributes the small cost of cache reuse to the similarity of keys and values across recursions~\citep{bae2025mixtureofrecursions}, as we see in \LT{}. With \MemorySharing{}, shared and local keys are instead clearly different. This shows that representation similarity across layers is not a necessary condition for effective memory sharing. This also fits the freedom that final loss gives: recursions need not all produce prediction-ready states, so the first can specialize in building memory and the later ones in using it. This specialization may help explain the quality gains, similar to how \citet{monea2026statepredictionseparationhypothesis} show that Transformers benefit from assigning memory construction and next-token prediction to separate residual streams.

\newtakeaway{Shared and local memory diverge in their local representations, so sharing is not reliant on recursions writing similar memories.}

\paragraph{Shared Memory Is a Gradient Highway}

Reading shared memory gives the final loss a direct gradient path to the first recursion, bypassing the intermediate recursions. We evaluate the \Std{}, \LT{}, and \LPT{} models of \autoref{tab:main-results} on the same $32$ validation sequences of $4{,}096$ tokens and backpropagate the final loss from $32$ fixed positions in each. We measure each recursion's contribution to the gradient on the Transformer block weights, and split the first recursion's contribution into its hidden-state and shared-KV paths by repeating the backward pass with shared KV detached (\autoref{app:analysis-details}).

\begin{figure}[!t]
  \centering
  \includegraphics[width=0.94\linewidth]{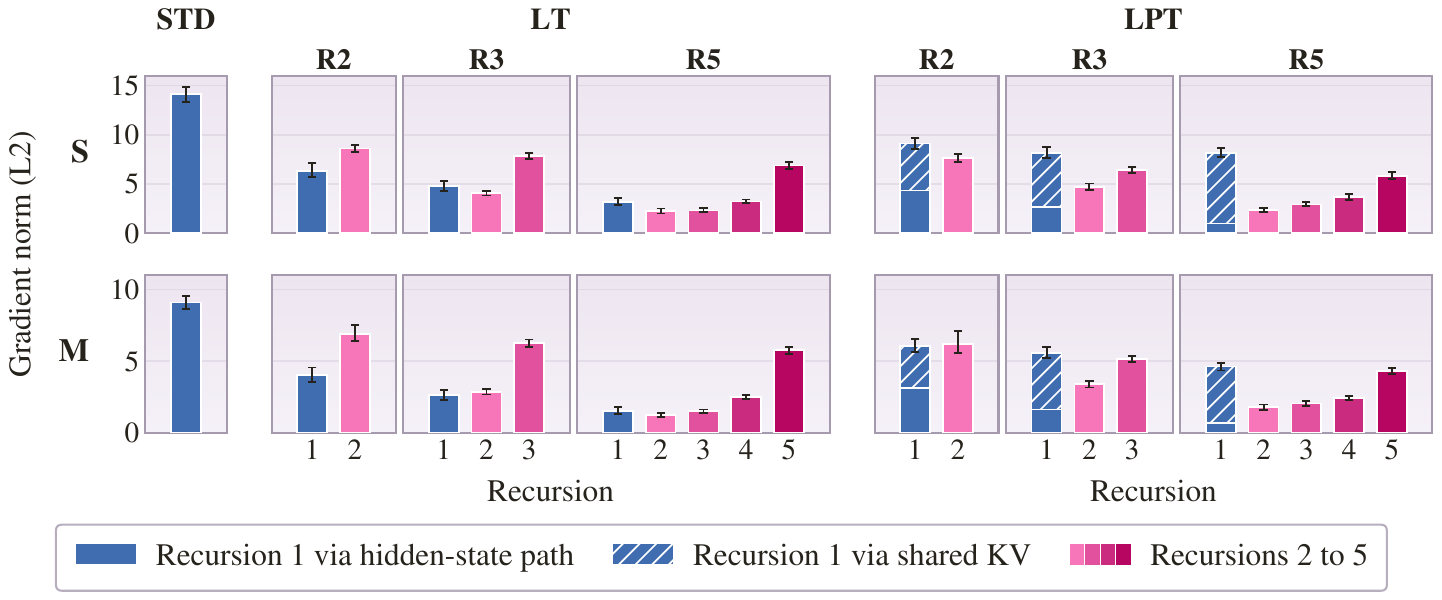}
  \caption{\textbf{Shared memory restores much of the gradient reaching the first recursion.} Mean gradient norm contributed by each recursion, with single-pass \Std{} as reference. The first \LPT{} bar is split into shared-KV (hatched) and hidden-state projections. \LT{} is trained with adaptive loss, the others with final loss. Whiskers are $95\%$ sequence-bootstrap intervals.}
  \label{fig:shared-feedback-recursions}
  \par\vspace{\floatsep}
  \setlength{\abovecaptionskip}{4pt}
  \includegraphics[width=\linewidth]{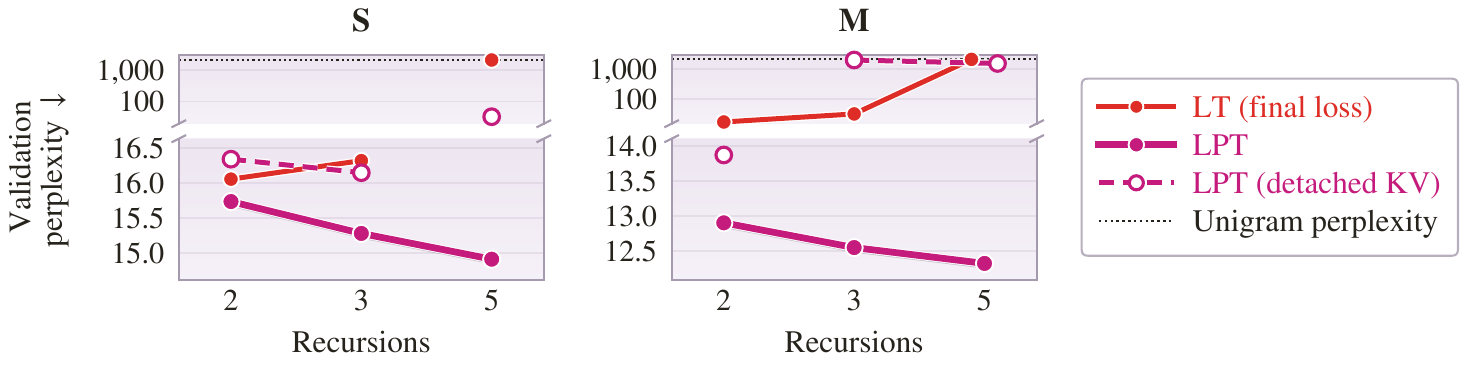}
  \caption{\textbf{Detaching shared-cache gradients hurts learning.} Validation perplexity ($\downarrow$), collapsed runs in the upper strip on a log scale. All models are trained with final loss; \autoref{tab:detach-ablation} lists the values.}
  \label{fig:detach-ablation}
\end{figure}

In \LT{}, the first-recursion gradient norm falls sharply with depth. \LPT{} recovers much of the signal magnitude relative to the single-pass \Std{} reference, with a much smaller decline as depth increases (\autoref{fig:shared-feedback-recursions}). At five recursions, most of this gradient comes through the shared KV cache. The same holds for \LPT{} with adaptive loss or full prediction caches (\autoref{fig:shared-feedback-conditions}).

To test whether this highway helps learning, we train \LPT{} from scratch with shared KV detached at later reads. The forward computation and hidden-state gradient path are unchanged. Perplexity rises in both sizes and at every depth, and deeper detached models often fail to train (\autoref{fig:detach-ablation}).

\newtakeaway{The memory sharing mechanism creates a gradient highway from the final loss to the first recursion that facilitates learning.}

\Needspace{6\baselineskip}
\section{Related Work}
\label{sec:related-work}

\paragraph{Separating Memory from Prediction}
In a standard Transformer, each hidden state does two jobs. It predicts the next token, and it writes the keys and values that later positions read. Several architectures separate these jobs across token streams, as in the two-stream attention of XLNet~\citep{yang2019xlnet}. The State-Prediction Separation Transformer~\citep{monea2026statepredictionseparationhypothesis} gives memory construction and next-token prediction separate residual streams and improves language modeling, and \citet{langford2026almostfree} study a variant of this split. A second family separates them across layers. The first layers write one cache once, and the remaining layers only read it~\citep{sun2024yoco,sun2026universalyoco,deepseekai2026deepseekv41flash,capps2026cart}. Instead, \LPT{} decouples memory from prediction across the recursions of a looped Transformer: the first recursion writes the model's only full-context memory, and every later recursion reads it while it predicts, keeping just a short window of its own. Looped models thus offer a third place for the split, with no extra stream or layer, and our analysis finds evidence that the recursions learn these roles (\autoref{sec:analysis}).

\paragraph{Looped Transformers and Memory Sharing}
Looped Transformers apply the same layers several times~\citep{dehghani2019universal,geiping2025huginn,ouro2025}, and recent work differs in which layers repeat and how recursions connect~\citep{gao2026loopies,wang2026smelt,huang2026loopedmodels,mozer2026recirculation,wang2026fullbandwidth,deng2026lt2}. In standard formulations, each recursion writes its own KV cache, so context memory grows with depth~\citep{geiping2025huginn,ouro2025}. Sharing a cache across recursions, like sharing it across layers~\citep{brandon2024crosslayerattention}, has been studied as a way to save memory. Without retraining, Huginn~\citep{geiping2025huginn} reuses caches cyclically with little quality loss, and Ouro~\citep{ouro2025} can keep only the final or an averaged cache. Keeping only the first recursion's cache sharply reduces Ouro's quality, which suggests that shared memory has to be trained. Trained variants reuse the first recursion's keys and values at a slight cost in quality~\citep{bae2025mixtureofrecursions}, compute them once in a separate prelude~\citep{capps2026cart}, or learn gated or compressed caches~\citep{vendrell2026melt,oneill2026loopedlatentattention}. \PLT{}~\citep{wu2025parallel} has the closest design to ours, although it focuses on parallelizing recursions. It shares the first recursion's cache with gated local caches and runs recursions in parallel, while \LPT{} keeps them sequential and reads both caches in one softmax. We find that \LPT{} performs better. All of these treat sharing as a cost to contain. We instead treat it as an inductive bias that separates memory from prediction and show that it improves quality (\autoref{sec:results} and \autoref{sec:analysis}).

\section{Discussion}
\label{sec:discussion}

We investigate memory sharing as an architectural inductive bias for looped models, and find that it not only cuts context memory but also improves quality in both full attention and hybrid models.  Our \MemorySharing{} \LPT{} models achieve consistently better perplexity and downstream accuracy than their usual looped counterparts, while matching the KV cache memory of a standard non-looped model. Our analyses suggest that sharing helps recursions learn different roles. The first recursion builds memory that later recursions use to refine the prediction. Those reads also give the final loss a direct gradient path to the first recursion. The different representations of shared and local memory support this division of roles, and cutting this gradient path hurts learning. Shared memory changes how later recursions read information and how the first recursion learns to write it.

Overall, our findings suggest that shared memory is a useful inductive bias for looped models, helping them make better use of computation within a limited memory budget.

\paragraph{Limitations \& Future work}Like other looped Transformers, \LPT{} is not compute-optimal: it trades time for quality, but unlike \LT{}, its memory barely grows with depth. At five recursions it costs $4.3$--$4.8\times$ the decoding FLOPs of \Std{}, about the same as \LT{}, and it spends the same number of recursions on every token. Choosing that number per token is future work. All our comparisons are parameter-matched, including \DeepStd{}, and we do not train a standard Transformer matched in decoding FLOPs instead, which at five recursions would have several times the parameters. We train up to $1$B parameters on one corpus with $4{,}096$-token contexts and do not test long-context tasks or, given our limited pretraining, very advanced tasks. Our analyses use S and M models, since not every configuration they need was trained at L, and only the detach ablation looks at training dynamics. Both the stability and memory improvements of \LPT{} are promising for effectively scaling looped models to higher recursion counts. More sophisticated \MemorySharing{} approaches could also be explored in future work, such as increasing the number of state recursions or alternating state recursions and prediction recursions.

\section*{Acknowledgments}
We thank Aviv Ron, Brian Kingsbury, and David Cox for useful feedback on internal presentations of this work.

\clearpage
\bibliographystyle{iclr2027_conference}
\bibliography{references}

\clearpage
\appendix
\raggedbottom
\section{Architecture and Training Details}
\label{app:experimental-details}

\paragraph{Architecture}
\autoref{tab:model-configurations} lists the base model dimensions. The decoder-only backbone uses RMSNorm pre-normalization~\citep{zhang2019rms}, SwiGLU feed-forward networks~\citep{shazeer2020glu}, no biases in linear layers, and tied input and output embeddings. Full-attention layers use rotary positional embeddings~\citep{su2021roformer}.

\begin{table}[!ht]
  \centering
  \setlength{\abovecaptionskip}{0pt}\setlength{\belowcaptionskip}{6pt}
  \caption{\textbf{Base model configurations.} Looped models reuse the listed $L$-layer stack.}
  \label{tab:model-configurations}
  \small
  \setlength{\tabcolsep}{10pt}
  \begin{tabular}{@{}lrrr@{}}
    \toprule
    & \textbf{S} & \textbf{M} & \textbf{L} \\
    \midrule
    Layers ($L$)      & 12     & 24     & 36 \\
    Hidden size ($d$) & 768    & 1,024  & 1,280 \\
    Heads             & 12     & 16     & 20 \\
    FFN size          & 3,072  & 4,096  & 5,120 \\
    Parameters        & 151.9M & 454.2M & 1.01B \\
    \bottomrule
  \end{tabular}
\end{table}

\LT{} and \LPT{} use Ouro's normalization and recurrent transition~\citep{ouro2025}. \LTtwo{} and \HLPT{} use the Full+GDN layer pattern and gating setup of \citet{deng2026lt2}, with one full-attention layer per five layers. The full-attention layers read the complete causal context in \LTtwo{}. In \HLPT{}, later recursions read the first recursion's full-context cache and their own local prediction window, as in \autoref{eq:lpt-attention}.

\PLT{} uses the same block, recurrent, and readout normalization as \LT{} and \LPT{}. Its recurrent transition adds the input embedding to the previous recursion's normalized output shifted by one token.

\paragraph{Attention Implementation}
During training, each prediction recursion concatenates its keys and values with the shared keys and values along the sequence dimension, separately at each full-attention layer. A block mask implements \autoref{eq:lpt-attention}, allowing full causal attention over shared entries and causal sliding-window attention over local entries. FlexAttention~\citep{dong2024flexattentionprogrammingmodel} skips fully masked blocks.

\paragraph{Data Processing}
We tokenize with the GPT-2 tokenizer~\citep{radford2019gpt2} and pack documents into $4{,}096$-token sequences, inserting an end-of-sequence token between documents. Attention is masked across document boundaries, and boundary tokens are excluded from the language-modeling loss.

\paragraph{Optimization}
We use AdamW with $\beta_1=0.9$, $\beta_2=0.95$, weight decay $0.1$, and gradient clipping at $1.0$. The global batch contains $96$ sequences of $4{,}096$ tokens, and training uses bfloat16 mixed precision. For L-size models, AdamW optimizer state is sharded across DDP ranks (ZeRO-1). The learning rate increases linearly to $6\times10^{-4}$ over the first $50$M tokens, stays constant, then decreases linearly to $6\times10^{-5}$ over the final $10\%$ of training. These are nanoGPT's default optimizer settings, used unchanged for every model, size, and objective: we did not tune the learning rate or the warmup for any model, and we did not test whether a lower peak learning rate or a longer warmup changes how final-loss \LT{} trains at depth (\autoref{fig:training-ablations}).

For adaptive loss, we use Ouro's adaptive-exit objective~\citep{ouro2025}. A learned per-token halting distribution weights the losses across recursions, with a KL regularization weight of $0.1$. Final loss applies the language-modeling loss only at the last recursion.

\section{Seed Variance and Statistical Tests}
\label{app:seeds}

Retraining every configuration at every size with several seeds is prohibitive at pretraining cost, so we run the check at S. Each S-size model of \autoref{tab:main-results} is trained with four seeds: the run reported in the table and three more, each changing the initialization and the data order together. All other settings stay at their main-table values. \autoref{tab:seeds-cells} lists the mean and $95\%$ confidence interval (Student-$t$) of the final validation perplexity for every model and recursion count. \autoref{tab:seeds-tests} tests each gap of interest with a one-sided paired $t$-test on the same-seed differences, with the alternative that the first model has lower perplexity. The largest standard deviation of perplexity across seeds is $0.08$, against gaps of $0.40$--$1.05$ between \LPT{} and \LT{}, and every difference keeps its sign at every seed.

\begin{table}[!ht]
  \centering
  \setlength{\abovecaptionskip}{0pt}\setlength{\belowcaptionskip}{6pt}
  \caption{\textbf{Seed variance at S.} Final validation perplexity, mean $\pm$ $95\%$ confidence interval (Student-$t$) over four seeds. In the \Std{} row, $\numrec=1$ is the standard Transformer and $\numrec>1$ the deeper, narrower parameter-matched Transformer. \SPS{} is the two-recursion baseline of \autoref{app:nine-recursions}.}
  \label{tab:seeds-cells}
  \small
  \setlength{\tabcolsep}{6pt}
  \begin{tabular}{@{}lllll@{}}
    \toprule
    Model & $\numrec=1$ & $\numrec=2$ & $\numrec=3$ & $\numrec=5$ \\
    \midrule
    \Std{} & $16.84 \pm 0.03$ & $17.03 \pm 0.05$ & $17.27 \pm 0.07$ & $17.64 \pm 0.13$ \\
    \SPS{} &  & $16.12 \pm 0.07$ &  &  \\
    \Loopie{} &  & $16.57 \pm 0.03$ & $16.57 \pm 0.08$ & $16.58 \pm 0.03$ \\
    \LT{} &  & $16.16 \pm 0.03$ & $16.00 \pm 0.05$ & $15.99 \pm 0.03$ \\
    \PLT{} &  & $15.83 \pm 0.05$ & $15.55 \pm 0.03$ & $15.46 \pm 0.03$ \\
    \LTtwo{} &  & $15.77 \pm 0.02$ & $15.47 \pm 0.02$ & $15.16 \pm 0.01$ \\
    \LPT{} &  & $15.76 \pm 0.07$ & $15.29 \pm 0.04$ & $14.94 \pm 0.04$ \\
    \HLPT{} &  & $15.73 \pm 0.02$ & $15.40 \pm 0.02$ & $15.03 \pm 0.02$ \\
    \bottomrule
  \end{tabular}
\end{table}

\begin{table}[!ht]
  \centering
  \setlength{\abovecaptionskip}{0pt}\setlength{\belowcaptionskip}{6pt}
  \caption{\textbf{Paired tests at S.} Same-seed perplexity gap of the first model over the second, mean $\pm$ $95\%$ confidence interval over four seeds, the gap in units of the standard deviation of the same-seed differences, and the one-sided paired $t$-test for the first model having lower perplexity.}
  \label{tab:seeds-tests}
  \small
  \setlength{\tabcolsep}{6pt}
  \begin{tabular}{@{}lcccc@{}}
    \toprule
    Comparison & $\numrec$ & Gap (PPL) & Gap$/$SD & $p$ \\
    \midrule
    \LPT{} vs.\ \LT{} & 2 & $-0.40 \pm 0.05$ & $13$ & $6.5\times10^{-5}$ \\
     & 3 & $-0.71 \pm 0.05$ & $24$ & $9.4\times10^{-6}$ \\
     & 5 & $-1.05 \pm 0.04$ & $39$ & $2.4\times10^{-6}$ \\
    \addlinespace[3pt]
    \HLPT{} vs.\ \LTtwo{} & 2 & $-0.04 \pm 0.01$ & $4$ & $1.7\times10^{-3}$ \\
     & 3 & $-0.07 \pm 0.02$ & $5$ & $9.8\times10^{-4}$ \\
     & 5 & $-0.13 \pm 0.02$ & $9$ & $1.6\times10^{-4}$ \\
    \addlinespace[3pt]
    \LPT{} vs.\ \PLT{} & 2 & $-0.07 \pm 0.09$ & $1$ & $4.8\times10^{-2}$ \\
     & 3 & $-0.27 \pm 0.07$ & $6$ & $5.4\times10^{-4}$ \\
     & 5 & $-0.52 \pm 0.07$ & $11$ & $9.7\times10^{-5}$ \\
    \bottomrule
  \end{tabular}
\end{table}

\section{Normalization between Recursions}
\label{app:lt-normalization}

Our \LT{} and \LPT{} use the Ouro paper's normalization~\citep{ouro2025}. Sandwich norms normalize the output of every sublayer before the residual addition, and the recurrent norm $\mathcal{N}_{\mathrm{rec}}$ normalizes each recursion's output before the next recursion (\autoref{eq:lt-recurrence}). With final loss, this \LT{} degrades or diverges at depth (\autoref{fig:training-ablations}), so the main tables train it with adaptive loss. We did not tune the normalization of either model. To test whether Ouro's normalization holds \LT{} back, we also train S and M \LT{} with final loss and five other normalizations, each shared across recursions:
\begin{itemize}
  \item \emph{Pre-norm only}~\citep{xiong2020layer} removes both, so $h_{\mathrm{in}}^{(t+1)}=h^{(t)}$ and each block keeps only its pre-norms.
  \item \emph{Sandwich only} keeps the sandwich norms~\citep{ding2021cogview} and removes $\mathcal{N}_{\mathrm{rec}}$.
  \item \emph{Input norm} removes the sandwich norms and keeps $\mathcal{N}_{\mathrm{rec}}$, which it also applies to the embeddings.
  \item \emph{Update norm} removes the sandwich norms and normalizes only each recursion's update, so the carried state stays unnormalized: $h_{\mathrm{in}}^{(t+1)}=h_{\mathrm{in}}^{(t)}+\mathcal{N}\big(h^{(t)}-h_{\mathrm{in}}^{(t)}\big)$.
  \item \emph{KEEL} removes the sandwich norms and applies the highway residual of \citet{chen2026keel} across recursions: $h_{\mathrm{in}}^{(t+1)}=\mathcal{N}_{2}\big(\numrec\,h_{\mathrm{in}}^{(t)}+\stack^{L}(\mathcal{N}_{1}(h_{\mathrm{in}}^{(t)}))\big)$.
\end{itemize}
Here $\mathcal{N}$, $\mathcal{N}_1$, and $\mathcal{N}_2$ are RMSNorms. Data, seed, batch, and learning-rate schedule match the main runs. \LPT{} keeps Ouro's normalization and final loss.

\begin{table}[!ht]
  \centering
  \setlength{\abovecaptionskip}{0pt}\setlength{\belowcaptionskip}{6pt}
  \caption{\textbf{\LPT{} outperforms \LT{} with the best normalization we tried.} Validation perplexity ($\downarrow$) after $20$B tokens. \textbf{Bold} marks the best per column and \underline{underline} the best \LT{}. Values above $2{,}000$ are unigram-level. A dash marks a run that did not finish.}
  \label{tab:lt-normalization}
  \small
  \setlength{\tabcolsep}{5pt}
  \begin{tabular}{@{}lll ccc ccc@{}}
    \toprule
    & & & \multicolumn{3}{c}{S} & \multicolumn{3}{c}{M} \\
    \cmidrule(lr){4-6}\cmidrule(lr){7-9}
    Model & Normalization & Loss & $\numrec=2$ & $\numrec=3$ & $\numrec=5$ & $\numrec=2$ & $\numrec=3$ & $\numrec=5$ \\
    \midrule
    \LT{} & Ouro (main) & Adaptive & 16.17 & 15.98 & 16.00 & 13.28 & 13.13 & 13.34 \\
     & Ouro & Final & 16.05 & 16.32 & 2{,}040 & 17.10 & 31.74 & 2{,}090 \\
     & Pre-norm only & Final & 16.28 & 15.99 & 15.77 & 13.28 & 13.05 & 12.83 \\
     & Sandwich only & Final & 16.23 & 15.83 & 15.67 & -- & -- & -- \\
     & Input norm & Final & 16.10 & 15.79 & 15.49 & 13.05 & 12.88 & 13.02 \\
     & Update norm & Final & {\underline{16.04}} & {\underline{15.65}} & 15.35 & 13.01 & {\underline{12.75}} & 12.77 \\
     & KEEL & Final & 16.10 & 15.73 & {\underline{15.24}} & {\underline{12.96}} & 12.78 & {\underline{12.76}} \\
    \midrule
    \rowcolor{lptrow}
    \textbf{\LPT{}} & Ouro & Final & {\bfseries 15.73} & {\bfseries 15.28} & {\bfseries 14.91} & {\bfseries 12.91} & {\bfseries 12.55} & {\bfseries 12.33} \\
    \bottomrule
  \end{tabular}
\end{table}

At every size and depth in \autoref{tab:lt-normalization}, \LPT{} with Ouro's normalization reaches lower perplexity than the best \LT{}, and the best \LT{} reaches lower perplexity than our main \LT{}. The alternatives are also more robust: every finished run trains with final loss at every size and depth, while Ouro's normalization degrades or diverges. No single normalization is best for \LT{}: update norm wins three settings and KEEL the other three. Choosing the best one lowers \LT{}'s perplexity by $0.13$--$0.76$ relative to our main \LT{}. \LPT{} still leads the best \LT{} by $0.31$--$0.37$ at S and by $0.06$--$0.43$ at M. The smallest gap, $0.06$ at M with two recursions, is within the seed variation we measure at S (\autoref{app:seeds}). At S, every alternative keeps improving through five recursions, unlike our main \LT{}. At M, the best \LT{} stops improving after three recursions.

Although this comparison favors \LT{}, \LPT{} still performs best. \LT{} gets the best of six normalizations, chosen on the same validation perplexity we report, while \LPT{} keeps Ouro's. The same search for \LPT{} would include Ouro's normalization, so it could only keep or lower \LPT{}'s perplexity. Since normalization is not the focus of our investigation, we keep Ouro's for the main \LT{}, as in the original paper.

\section{Additional Analysis Details}
\label{app:analysis-details}

\paragraph{Attention Representations}
\label{app:attention-analysis}
For \autoref{fig:qkv-pca}, we use S-size models with three recursions and final loss. We evaluate each model on the same $4{,}096$-token validation sequence. At each layer, we concatenate the attention heads into one query vector and one key vector per token, before rotary encoding. We pool these vectors across all tokens and recursions, subtract their common mean, and fit PCA by singular value decomposition, separately for each model and layer. We do not normalize the vectors or rescale their features. All recursions within a panel therefore share one basis. Axes are not aligned across models or layers.

We display the first two principal components. For readability, the panels show one token in $24$, while PCA and the large centroid markers use all tokens. Lines connect centroids across recursions, separately for queries and keys. Axis limits are shared across layers within each model, with equal scaling on both axes.

\begin{figure}[!t]
  \centering
  \includegraphics[width=\linewidth]{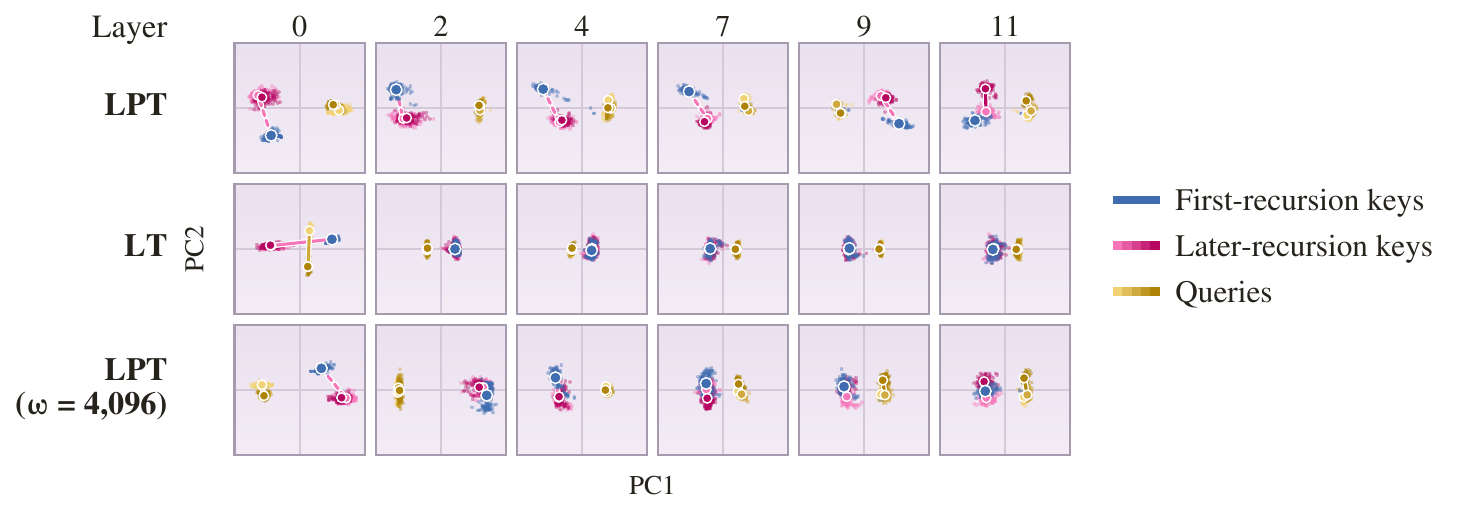}
  \caption{\textbf{Shared and local keys remain distinguishable with full prediction caches.} Expanded comparison for \autoref{fig:qkv-pca}, including $\wpred=4{,}096$.}
  \label{fig:qkv-pca-full}
\end{figure}

\begin{figure}[!t]
  \centering
  \includegraphics[width=\linewidth]{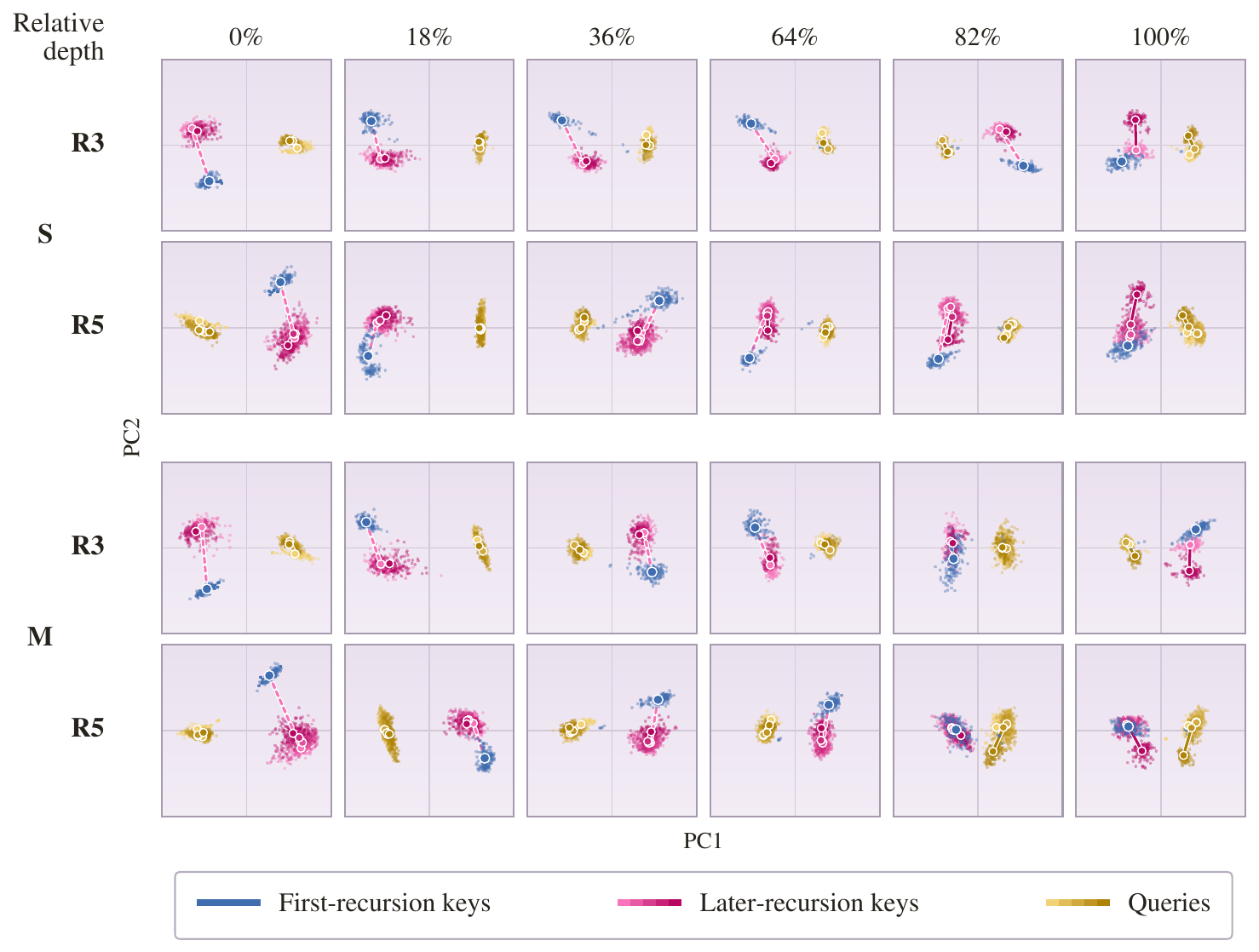}
  \caption{\textbf{Query/key geometry across \LPT{} models.} S and M models at three and five recursions, with final loss and $\wpred=64$. Columns match relative layer depth.}
  \label{fig:lpt-pca-depth}
\end{figure}

\begin{figure}[!t]
  \centering
  \includegraphics[width=\linewidth]{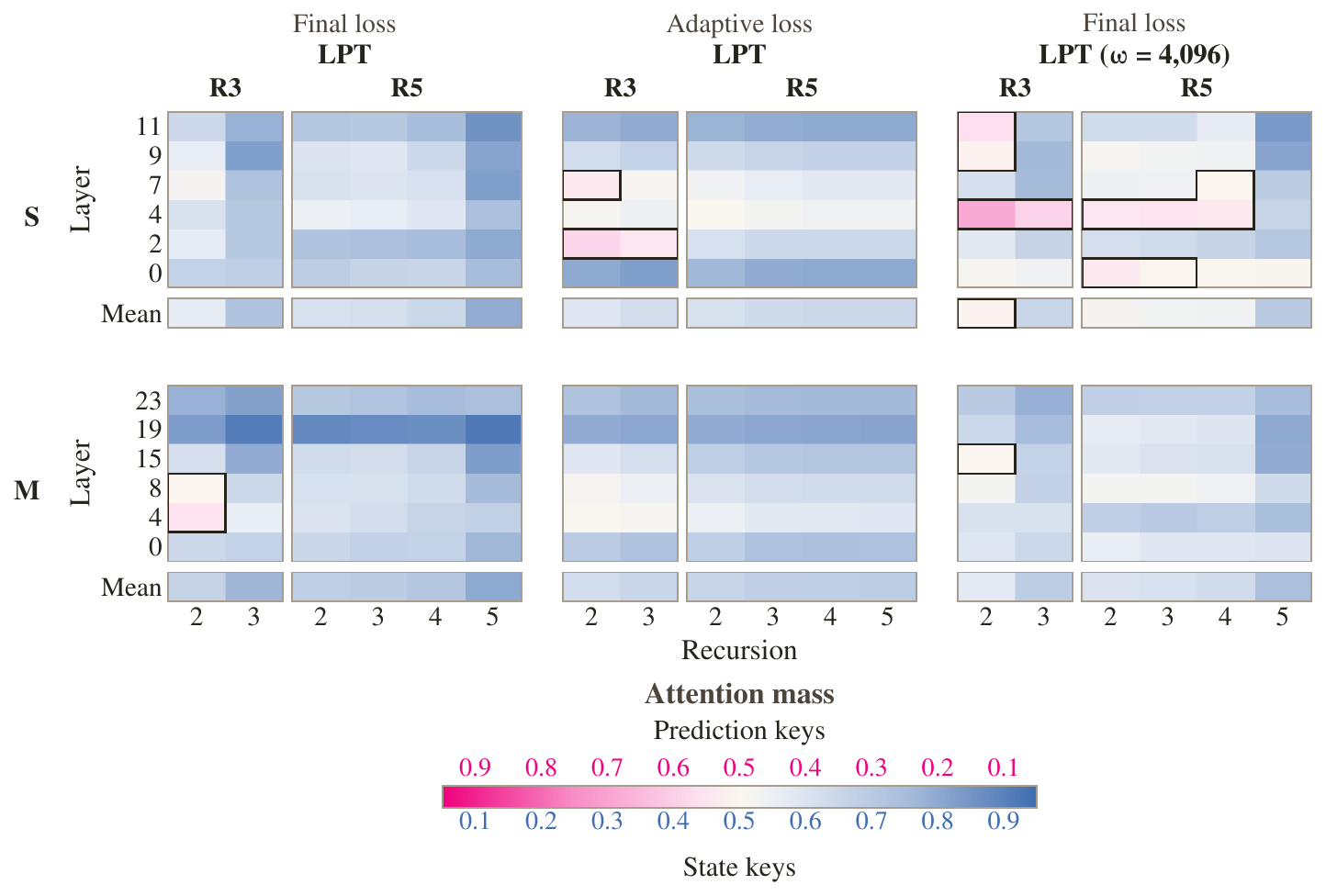}
  \caption{\textbf{Attention to shared memory across training objectives and window sizes.} Expanded comparison for \autoref{fig:state-attention-mass}. Mean rows average all layers.}
  \label{fig:state-attention-mass-conditions}
\end{figure}

\autoref{fig:qkv-pca-full} adds \LPT{} with full prediction caches. \autoref{fig:lpt-pca-depth} compares \LPT{} across sizes and recursion counts, matching layers by relative depth. \autoref{fig:state-attention-mass-conditions} compares attention to shared memory across window sizes and training objectives. Each layer's attention mass is averaged over queries and heads. We display six layers at matched relative depths, but compute the mean over all $12$ layers for S and all $24$ for M.

\begin{figure}[!t]
  \centering
  \includegraphics[width=\linewidth]{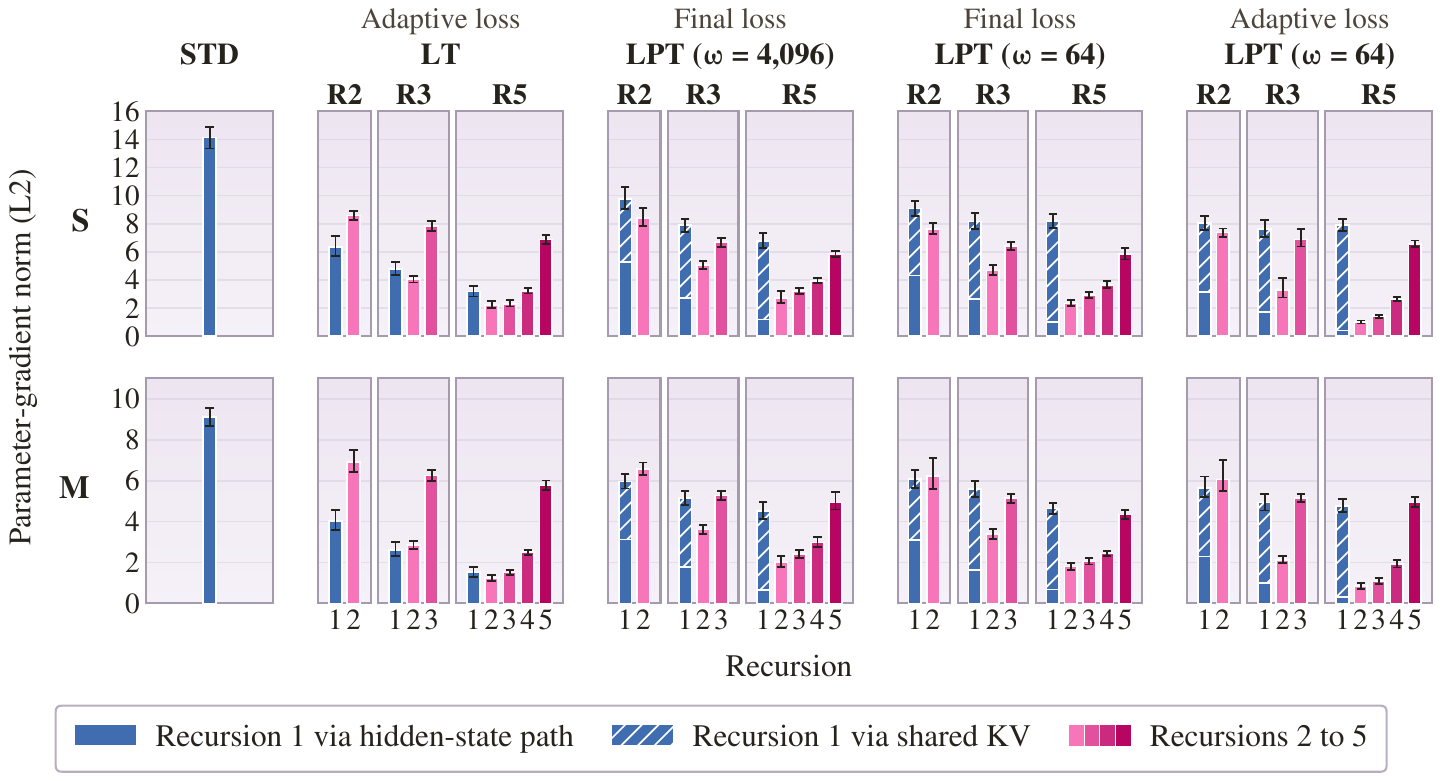}
  \caption{\textbf{Shared-cache feedback persists across training objectives and window sizes.} Expanded version of \autoref{fig:shared-feedback-recursions}, with the same sequences and final-loss gradient probe in every condition.}
  \label{fig:shared-feedback-conditions}
\end{figure}

\paragraph{Gradient Feedback through Shared Memory}
\label{app:gradient-feedback}
We differentiate the mean cross-entropy of the final recursion at $32$ sampled positions in each of $32$ validation sequences of length $4{,}096$. The sequence contents and loss positions are identical across conditions. \autoref{fig:shared-feedback-recursions} uses the main-comparison checkpoints: \LT{} trained with adaptive loss and \LPT{} with final loss. \Std{} provides a single-pass reference, measured through its native forward pass on the same data. \autoref{fig:shared-feedback-conditions} adds \LPT{} with full prediction caches and \LPT{} trained with adaptive loss. Every probe differentiates the loss at the final recursion, regardless of the training objective. We measure gradients on Transformer block parameters, excluding embeddings, readout, recurrent normalization, and the halting gate. Each model retains its trained normalization.

For the backward computation, we give each recursion an independent parameter copy with the same values. This leaves the forward pass unchanged and isolates each recursion's contribution to the shared-parameter gradient. Detaching shared KV at later reads isolates the first recursion's hidden-state path contribution, $g_{\mathrm{hidden}}$. Subtracting it from the ordinary first-recursion gradient $g^{(1)}$ gives the shared-cache contribution, $g_{\mathrm{KV}}$. We verify that contributions from all parameter uses sum to the ordinary shared-parameter gradient.

The bars show arithmetic mean gradient norms across sequences. To split the first bar, we project each path onto its total gradient within each sequence:
\begin{equation}
  p_a = \frac{\langle g_a,g^{(1)}\rangle}{\lVert g^{(1)}\rVert_2},
  \qquad a\in\{\mathrm{KV},\mathrm{hidden}\},
  \qquad p_{\mathrm{KV}}+p_{\mathrm{hidden}}=\lVert g^{(1)}\rVert_2.
  \label{eq:gradient-projection}
\end{equation}
We then average these projected contributions. Their sum equals the mean first-recursion norm. The two path norms themselves are not additive. The projections are nonnegative in all plotted measurements, without clipping. Whiskers are $95\%$ sequence-bootstrap intervals for the full bar, not for its segments or for variation across training seeds.

\autoref{fig:shared-feedback-conditions} shows that the shared-cache contribution remains strong with a full prediction window and under adaptive loss. This is consistent across S and M. The compact main-paper comparison therefore does not depend on using final loss or a short prediction window for \LPT{}.

\begin{figure}[!t]
  \centering
  \includegraphics[width=\linewidth]{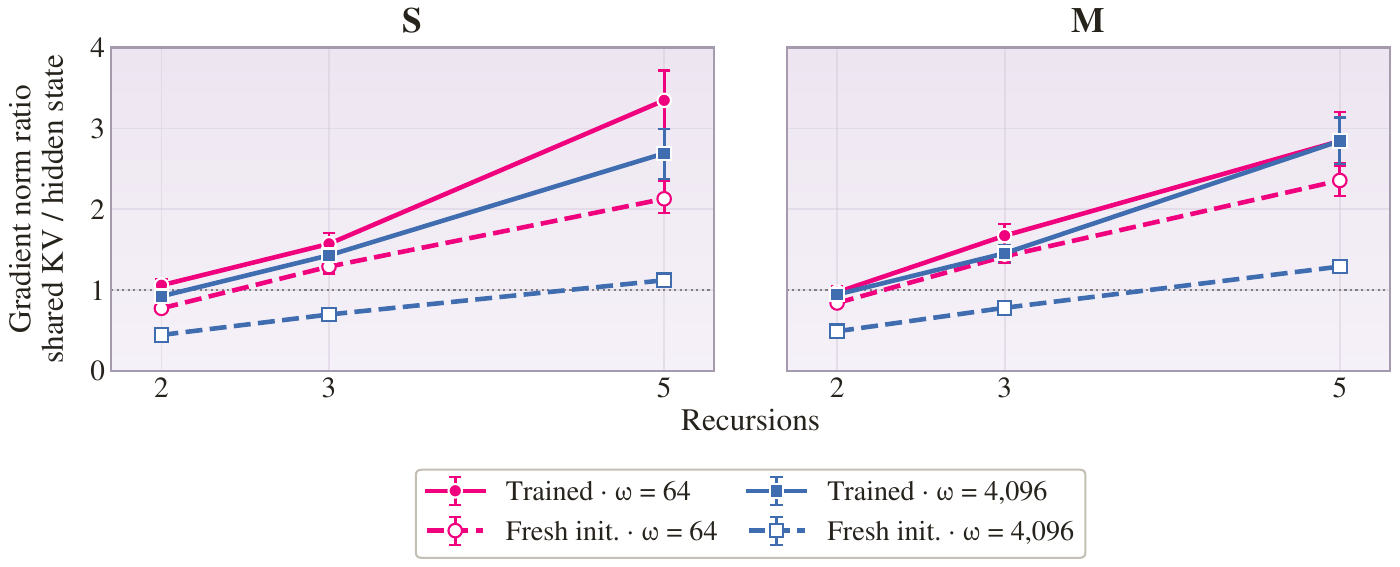}
  \caption{\textbf{Shared-KV feedback relative to the hidden-state path.} Gradient-norm ratios at fresh initialization and after training. Whiskers: $95\%$ sequence-bootstrap intervals.}
  \label{fig:shared-feedback-depth}
\end{figure}

\paragraph{Gradient Paths at Initialization}
\autoref{fig:shared-feedback-depth} compares the norm ratio $\lVert g_{\mathrm{KV}}\rVert_2/\lVert g_{\mathrm{hidden}}\rVert_2$ before and after training, using geometric means over the same $32$ sequences. Fresh initialization is measured at two, three, and five recursions for both windows and sizes. It uses separate random weights, not saved initial checkpoints. The ratio increases with depth even at initialization, but is larger in the trained models. More recursions add more shared-cache reads, so this trend alone does not establish that the feedback helps learning. The detachment experiment tests that separately.

\paragraph{Shared-Cache Gradient Ablation}
We train S and M \LPT{} models with $\numrec\in\{2,3,5\}$, $\wpred=64$, and final loss, changing only whether the shared KV entries are detached at each read by a later recursion. The first recursion still receives gradients through the hidden-state path, and all shared parameters remain trainable. This removes direct gradient feedback through shared-cache reads, not all training of the first recursion.

\autoref{tab:detach-ablation} compares final-loss \LT{}, detached \LPT{}, and \LPT{} after $20$B training tokens. Detachment raises perplexity in every comparison. Detached S at five recursions finishes well above \LPT{}, while M at three and five recursions remains near unigram-level loss. The experiment supports a role for shared-cache gradients in learning and stability, but does not distinguish shorter gradient paths from changes in gradient scale or direction.

\begin{table}[!ht]
  \centering
  \setlength{\abovecaptionskip}{0pt}\setlength{\belowcaptionskip}{6pt}
  \caption{\textbf{Detaching shared-cache gradients hurts learning.} Validation perplexity ($\downarrow$) at S and M, all models trained with final loss; values above $1{,}000$ are runs that failed to train. Bold marks the best per column.}
  \label{tab:detach-ablation}
  \small
  \setlength{\tabcolsep}{3pt}
  \renewcommand{\arraystretch}{1.08}
  \begin{tabular*}{\linewidth}{@{\extracolsep{\fill}}l rrr rrr@{}}
    \toprule
    & \multicolumn{3}{c}{S} & \multicolumn{3}{c}{M} \\
    \cmidrule(lr){2-4}\cmidrule(lr){5-7}
    Model & $\numrec=2$ & $\numrec=3$ & $\numrec=5$ & $\numrec=2$ & $\numrec=3$ & $\numrec=5$ \\
    \midrule
    \LT{} (final loss) & 16.05 & 16.32 & 2{,}040 & 17.10 & 31.74 & 2{,}090 \\
    \LPT{} (detached KV) & 16.34 & 16.14 & 33.21 & 13.87 & 1{,}995 & 1{,}527 \\
    \textbf{\LPT{}} & \textbf{15.73} & \textbf{15.28} & \textbf{14.91} & \textbf{12.91} & \textbf{12.55} & \textbf{12.33} \\
    \bottomrule
  \end{tabular*}
\end{table}

\section{Experiments at Nine Recursions}
\label{app:nine-recursions}
\label{app:frontier}

\begin{figure}[H]
  \centering
  \includegraphics[width=\linewidth]{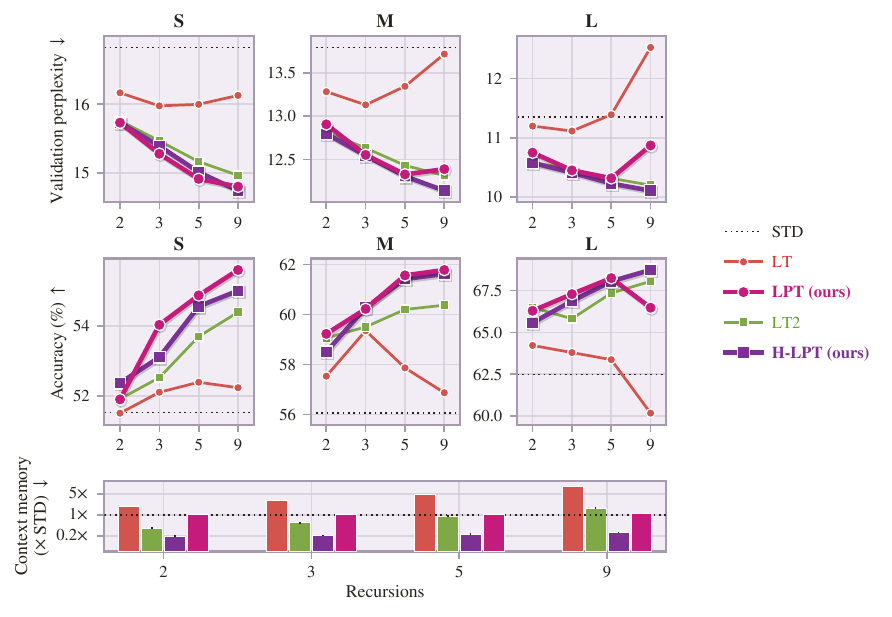}
  \caption{\textbf{Quality and memory through nine recursions.} Validation perplexity, mean five-task accuracy, and context memory relative to \Std{}.}
  \label{fig:acc-vs-recursions}
\end{figure}

\autoref{tab:main-results-full} and \autoref{fig:acc-vs-recursions} extend the comparisons to nine recursions. We also list \SPS{}~\citep{monea2026statepredictionseparationhypothesis} in the $\numrec=2$ column, since its memory and FLOPs match those of \LPT{} with two recursions. From five to nine, \LPT{} gains $0.7$ accuracy points at S and $0.2$ at M, but loses $1.7$ at L. Perplexity improves at S and worsens at M and L. \LT{} and \PLT{} also worsen in perplexity at every size. Both hybrids continue to improve: \HLPT{} gains $0.4$, $0.2$, and $0.6$ accuracy points at S, M, and L, respectively, and reaches the lowest perplexity of all models at every size. At nine recursions, \LPT{} still outperforms \LT{} at every size, and \HLPT{} outperforms \LTtwo{} in both perplexity and accuracy.

Higher recurrent depth can also complicate training. Ouro reports loss spikes and gradient oscillations at eight recursions and reduces its recurrent depth to four~\citep{ouro2025}. \LTtwo{} also uses four recursions~\citep{deng2026lt2}.

\begin{table}[H]
    \centering
    \setlength{\abovecaptionskip}{0pt}\setlength{\belowcaptionskip}{6pt}
    \caption{\textbf{Quality and efficiency through nine recursions.} One table per size. Metrics and formatting follow \autoref{tab:main-results}: in the \Std{} row, $\numrec=1$ is the standard Transformer and $\numrec>1$ is \DeepStd{}. \SPS{} appears in the $\numrec=2$ column, whose memory and FLOPs it matches.}
    \label{tab:main-results-full}
    \scriptsize
    \setlength{\tabcolsep}{2pt}
    \renewcommand{\arraystretch}{1.15}
    \resizebox{\textwidth}{!}{%
    \begin{tabular}{@{}l S[table-format=2.2] S[table-format=2.1] S[table-format=2.2] S[table-format=2.1] S[table-format=1.2] S[table-format=1.2] S[table-format=2.2] S[table-format=2.1] S[table-format=1.2] S[table-format=1.2] S[table-format=2.2] S[table-format=2.1] S[table-format=1.2] S[table-format=1.2] S[table-format=2.2] S[table-format=2.1] S[table-format=1.2] S[table-format=1.2]@{}}
        \toprule
        \textbf{S} & \multicolumn{2}{c}{$\numrec=1$} & \multicolumn{4}{c}{$\numrec=2$} & \multicolumn{4}{c}{$\numrec=3$} & \multicolumn{4}{c}{$\numrec=5$} & \multicolumn{4}{c}{$\numrec=9$} \\
        \cmidrule(lr){2-3} \cmidrule(lr){4-7} \cmidrule(lr){8-11} \cmidrule(lr){12-15} \cmidrule(lr){16-19}
        Model & {PPL $\downarrow$} & {Acc. $\uparrow$} & {PPL $\downarrow$} & {Acc. $\uparrow$} & {KV $\downarrow$} & {FLOPs $\downarrow$} & {PPL $\downarrow$} & {Acc. $\uparrow$} & {KV $\downarrow$} & {FLOPs $\downarrow$} & {PPL $\downarrow$} & {Acc. $\uparrow$} & {KV $\downarrow$} & {FLOPs $\downarrow$} & {PPL $\downarrow$} & {Acc. $\uparrow$} & {KV $\downarrow$} & {FLOPs $\downarrow$} \\
        \midrule
        \Std{} & 16.83 & 51.5 & 17.07 & 50.2 & 1.50 & {\bfseries 1.19} & 17.27 & 50.7 & 1.88 & {\bfseries 1.31} & 17.69 & 50.4 & 2.50 & {\bfseries 1.54} & 18.32 & 49.1 & 3.38 & {\bfseries 1.81} \\
        \SPS{} &  &  & 16.18 & {\underline{52.6}} & 1.02 & 1.84 &  &  &  &  &  &  &  &  &  &  &  &  \\
        \Loopie{} &  &  & 16.58 & 51.6 & 2.00 & 1.83 & 16.50 & 51.9 & 3.00 & 2.66 & 16.56 & 52.1 & 5.00 & 4.32 & 16.98 & 51.0 & 9.00 & 7.64 \\
        \LT{} &  &  & 16.17 & 51.5 & 2.00 & 1.83 & 15.98 & 52.1 & 3.00 & 2.66 & 16.00 & 52.4 & 5.00 & 4.32 & 16.13 & 52.2 & 9.00 & 7.64 \\
        \PLT{} &  &  & 15.84 & {\bfseries 52.7} & 1.02 & 1.84 & 15.55 & 52.9 & 1.03 & 2.67 & 15.49 & 53.5 & 1.06 & 4.34 & 15.51 & 53.2 & {\underline{1.12}} & 7.68 \\
        \LTtwo{} &  &  & {\underline{15.76}} & 51.9 & {\underline{0.35}} & {\underline{1.29}} & 15.47 & 52.5 & {\underline{0.52}} & {\underline{1.86}} & 15.16 & 53.7 & {\underline{0.87}} & {\underline{2.98}} & 14.96 & 54.4 & 1.56 & {\underline{5.23}} \\
        \rowcolor{lptrow}
        \textbf{\LPT{}} &  &  & {\bfseries 15.73} & 51.9 & 1.02 & 1.84 & {\bfseries 15.28} & {\bfseries 54.0} & 1.03 & 2.67 & {\bfseries 14.91} & {\bfseries 54.9} & 1.06 & 4.34 & {\underline{14.80}} & {\bfseries 55.6} & {\underline{1.12}} & 7.68 \\
        \rowcolor{lptrow}
        \textbf{\HLPT{}} &  &  & {\bfseries 15.73} & 52.4 & {\bfseries 0.18} & {\underline{1.29}} & {\underline{15.39}} & {\underline{53.1}} & {\bfseries 0.19} & {\underline{1.86}} & {\underline{15.01}} & {\underline{54.6}} & {\bfseries 0.21} & {\underline{2.98}} & {\bfseries 14.74} & {\underline{55.0}} & {\bfseries 0.25} & {\underline{5.23}} \\
        \bottomrule
    \end{tabular}%
    }
    \par\vspace{7pt}
    \resizebox{\textwidth}{!}{%
    \begin{tabular}{@{}l S[table-format=2.2] S[table-format=2.1] S[table-format=2.2] S[table-format=2.1] S[table-format=1.2] S[table-format=1.2] S[table-format=2.2] S[table-format=2.1] S[table-format=1.2] S[table-format=1.2] S[table-format=2.2] S[table-format=2.1] S[table-format=1.2] S[table-format=1.2] S[table-format=2.2] S[table-format=2.1] S[table-format=1.2] S[table-format=1.2]@{}}
        \toprule
        \textbf{M} & \multicolumn{2}{c}{$\numrec=1$} & \multicolumn{4}{c}{$\numrec=2$} & \multicolumn{4}{c}{$\numrec=3$} & \multicolumn{4}{c}{$\numrec=5$} & \multicolumn{4}{c}{$\numrec=9$} \\
        \cmidrule(lr){2-3} \cmidrule(lr){4-7} \cmidrule(lr){8-11} \cmidrule(lr){12-15} \cmidrule(lr){16-19}
        Model & {PPL $\downarrow$} & {Acc. $\uparrow$} & {PPL $\downarrow$} & {Acc. $\uparrow$} & {KV $\downarrow$} & {FLOPs $\downarrow$} & {PPL $\downarrow$} & {Acc. $\uparrow$} & {KV $\downarrow$} & {FLOPs $\downarrow$} & {PPL $\downarrow$} & {Acc. $\uparrow$} & {KV $\downarrow$} & {FLOPs $\downarrow$} & {PPL $\downarrow$} & {Acc. $\uparrow$} & {KV $\downarrow$} & {FLOPs $\downarrow$} \\
        \midrule
        \Std{} & 13.79 & 56.1 & 14.24 & 55.5 & 1.44 & {\bfseries 1.13} & 14.47 & 55.4 & 1.78 & {\bfseries 1.24} & 14.88 & 54.4 & 2.34 & {\bfseries 1.43} & 15.60 & 53.4 & 3.09 & {\bfseries 1.63} \\
        \SPS{} &  &  & 12.99 & 58.9 & 1.02 & 1.93 &  &  &  &  &  &  &  &  &  &  &  &  \\
        \Loopie{} &  &  & 13.69 & 56.9 & 2.00 & 1.92 & 13.75 & 55.8 & 3.00 & 2.84 & 13.74 & 56.6 & 5.00 & 4.69 & 15.80 & 52.9 & 9.00 & 8.37 \\
        \LT{} &  &  & 13.28 & 57.5 & 2.00 & 1.92 & 13.13 & 59.4 & 3.00 & 2.84 & 13.34 & 57.9 & 5.00 & 4.69 & 13.72 & 56.9 & 9.00 & 8.37 \\
        \PLT{} &  &  & 13.00 & {\underline{59.1}} & 1.02 & 1.93 & 12.68 & 60.0 & 1.03 & 2.85 & 12.65 & 60.5 & 1.06 & 4.71 & 12.87 & 59.5 & {\underline{1.12}} & 8.41 \\
        \LTtwo{} &  &  & {\underline{12.81}} & {\underline{59.1}} & {\underline{0.35}} & {\underline{1.43}} & 12.63 & 59.5 & {\underline{0.52}} & {\underline{2.10}} & 12.43 & 60.2 & {\underline{0.87}} & {\underline{3.45}} & {\underline{12.31}} & 60.4 & 1.56 & {\underline{6.14}} \\
        \rowcolor{lptrow}
        \textbf{\LPT{}} &  &  & 12.91 & {\bfseries 59.2} & 1.02 & 1.93 & {\underline{12.55}} & {\underline{60.2}} & 1.03 & 2.85 & {\underline{12.33}} & {\bfseries 61.6} & 1.06 & 4.70 & 12.39 & {\bfseries 61.8} & {\underline{1.12}} & 8.41 \\
        \rowcolor{lptrow}
        \textbf{\HLPT{}} &  &  & {\bfseries 12.80} & 58.5 & {\bfseries 0.18} & {\underline{1.43}} & {\bfseries 12.54} & {\bfseries 60.3} & {\bfseries 0.19} & {\underline{2.10}} & {\bfseries 12.31} & {\underline{61.4}} & {\bfseries 0.21} & {\underline{3.45}} & {\bfseries 12.14} & {\underline{61.6}} & {\bfseries 0.25} & {\underline{6.14}} \\
        \bottomrule
    \end{tabular}%
    }
    \par\vspace{7pt}
    \resizebox{\textwidth}{!}{%
    \begin{tabular}{@{}l S[table-format=2.2] S[table-format=2.1] S[table-format=2.2] S[table-format=2.1] S[table-format=1.2] S[table-format=1.2] S[table-format=2.2] S[table-format=2.1] S[table-format=1.2] S[table-format=1.2] S[table-format=2.2] S[table-format=2.1] S[table-format=1.2] S[table-format=1.2] S[table-format=2.2] S[table-format=2.1] S[table-format=1.2] S[table-format=1.2]@{}}
        \toprule
        \textbf{L} & \multicolumn{2}{c}{$\numrec=1$} & \multicolumn{4}{c}{$\numrec=2$} & \multicolumn{4}{c}{$\numrec=3$} & \multicolumn{4}{c}{$\numrec=5$} & \multicolumn{4}{c}{$\numrec=9$} \\
        \cmidrule(lr){2-3} \cmidrule(lr){4-7} \cmidrule(lr){8-11} \cmidrule(lr){12-15} \cmidrule(lr){16-19}
        Model & {PPL $\downarrow$} & {Acc. $\uparrow$} & {PPL $\downarrow$} & {Acc. $\uparrow$} & {KV $\downarrow$} & {FLOPs $\downarrow$} & {PPL $\downarrow$} & {Acc. $\uparrow$} & {KV $\downarrow$} & {FLOPs $\downarrow$} & {PPL $\downarrow$} & {Acc. $\uparrow$} & {KV $\downarrow$} & {FLOPs $\downarrow$} & {PPL $\downarrow$} & {Acc. $\uparrow$} & {KV $\downarrow$} & {FLOPs $\downarrow$} \\
        \midrule
        \Std{} & 11.35 & 62.5 & 11.62 & 62.5 & 1.45 & {\bfseries 1.14} & 11.96 & 61.3 & 1.73 & {\bfseries 1.17} & 12.27 & 61.4 & 2.25 & {\bfseries 1.32} & 12.66 & 59.8 & 3.15 & {\bfseries 1.63} \\
        \SPS{} &  &  & 10.77 & 65.5 & 1.02 & 1.96 &  &  &  &  &  &  &  &  &  &  &  &  \\
        \Loopie{} &  &  & 11.26 & 63.5 & 2.00 & 1.95 & 11.28 & 63.9 & 3.00 & 2.91 & 12.26 & 61.0 & 5.00 & 4.81 & 11.53 & 63.2 & 9.00 & 8.63 \\
        \LT{} &  &  & 11.20 & 64.2 & 2.00 & 1.95 & 11.11 & 63.8 & 3.00 & 2.91 & 11.39 & 63.4 & 5.00 & 4.81 & 12.52 & 60.2 & 9.00 & 8.63 \\
        \PLT{} &  &  & 10.95 & 64.8 & 1.02 & 1.96 & 10.75 & 66.8 & 1.03 & 2.92 & 10.59 & 66.4 & 1.06 & 4.83 & 10.78 & 66.1 & {\underline{1.12}} & 8.66 \\
        \LTtwo{} &  &  & {\underline{10.59}} & {\bfseries 66.5} & {\underline{0.40}} & {\underline{1.53}} & 10.47 & 65.8 & {\underline{0.60}} & {\underline{2.27}} & {\underline{10.32}} & 67.3 & {\underline{1.01}} & {\underline{3.75}} & {\underline{10.20}} & {\underline{68.1}} & 1.81 & {\underline{6.71}} \\
        \rowcolor{lptrow}
        \textbf{\LPT{}} &  &  & 10.75 & {\underline{66.3}} & 1.02 & 1.96 & {\underline{10.45}} & {\bfseries 67.3} & 1.03 & 2.92 & {\underline{10.32}} & {\bfseries 68.2} & 1.06 & 4.83 & 10.87 & 66.5 & {\underline{1.12}} & 8.66 \\
        \rowcolor{lptrow}
        \textbf{\HLPT{}} &  &  & {\bfseries 10.58} & 65.6 & {\bfseries 0.21} & {\underline{1.53}} & {\bfseries 10.41} & {\underline{66.9}} & {\bfseries 0.22} & {\underline{2.27}} & {\bfseries 10.23} & {\underline{68.1}} & {\bfseries 0.24} & {\underline{3.75}} & {\bfseries 10.11} & {\bfseries 68.7} & {\bfseries 0.28} & 6.72 \\
        \bottomrule
    \end{tabular}%
    }
\end{table}

\paragraph{Intermediate Readouts}
We apply the model's readout normalization and language-modeling head to the output of each recursion. The measurements use S-size \LPT{} checkpoints trained for $20$B tokens, averaged over $16$ validation sequences of $4{,}096$ tokens. With nine recursions and final loss, intermediate-readout perplexity ranges from $87$ to $1{,}012$, falling to $15.6$ at the last recursion. With adaptive loss, it is already $2.79$ by the fourth recursion and remains near that value. Under final loss, the head is trained only on the last recursion: high intermediate perplexity measures poor prediction through this head, not the absence of predictive information in the hidden state. These probe-set losses are separate from the full validation-set perplexity in \autoref{tab:main-results-full}.

\begin{figure}[H]
  \centering
  \includegraphics[width=0.90\linewidth]{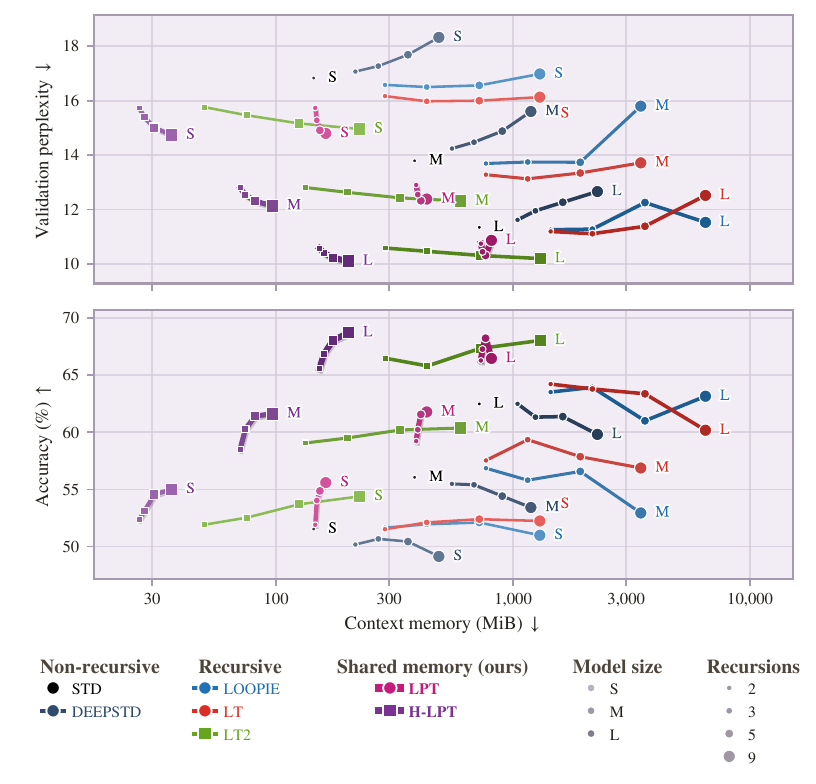}
  \caption{\textbf{Quality versus context memory across model sizes.} Curves vary recursion count. Marker size denotes recursions, and shade denotes model size. S/M train for $20$B tokens and L for $50$B.}
  \label{fig:frontier-allsizes}
\end{figure}

\begin{figure}[H]
  \centering
  \includegraphics[width=\linewidth]{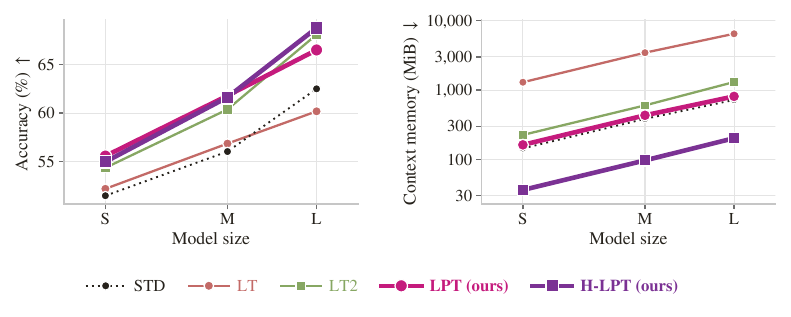}
  \caption{\textbf{Quality and memory across model sizes.} Mean five-task accuracy and context memory for \Std{} and looped models at nine recursions.}
  \label{fig:quality-vs-size}
\end{figure}

\paragraph{Complete Quality--Memory Sweep}
\autoref{fig:frontier-allsizes} includes every family, size, and recursion count in \autoref{tab:main-results-full} except \SPS{} and \PLT{}. \autoref{fig:quality-vs-size} compares nine-recursion models with \Std{} across sizes.

\section{Decoding Speed and Memory on a GPU}
\label{app:wallclock}

We test whether the analytic memory savings translate into larger batches and higher decoding throughput on a GPU. A smaller cache lets more sequences share a GPU, which can raise throughput~\citep{brandon2024crosslayerattention,sun2024yoco}. As \citet{oneill2026loopedlatentattention} do for a looped model, we measure the largest batch that fits at a fixed context and the throughput at that batch.

\paragraph{Setup}
We decode on one NVIDIA H100 with $80$ GB of memory, in bf16, at a $4{,}096$-token context: each step generates one token per sequence from a cache already filled to $4{,}068$ tokens. All models share one decoding implementation in PyTorch: a preallocated cache per attention layer, FlexAttention~\citep{dong2024flexattentionprogrammingmodel} for attention, and CUDA graphs that replay a whole decoding step, all recursions and layers included. A later recursion of \LPT{} or \HLPT{} attends to its shared and local caches in one softmax, which we compute by merging the two FlexAttention outputs by their log-sum-exp. Gated DeltaNet layers update their recurrent state one token at a time, with the kernels of flash-linear-attention~\citep{yang2024fla}. \Std{} runs the same code with one recursion, and \LT{} runs all its recursions, without early exit. Weights are random, since speed and memory do not depend on them. On the same GPU, decoding reproduces the full forward pass to bf16 precision. The maximum batch is the largest batch whose decoding steps run without running out of memory, and we measure throughput at this batch, or at $95\%$ of it when the CUDA graph does not fit. These measurements come from one research implementation, not an optimized serving engine, which would change the absolute speeds. Every model runs the same code, so the comparison holds within this implementation.

\paragraph{Capacity}
\autoref{tab:wallclock} shows that the analytic savings carry over to the GPU. \LPT{}'s cache per sequence is at most $6.25\%$ larger than \Std{}'s, so it fits $93$--$98\%$ of \Std{}'s batch. \LT{}'s cache grows with depth, so \LPT{} fits $2.0$, $2.9$, and $4.7$--$4.8$ times as many sequences as \LT{} at two, three, and five recursions. \HLPT{} holds $20$--$27\%$ of \Std{}'s cache and fits $3.6$--$5.0$ times as many sequences as \Std{}, and $1.8$--$3.8$ times as many as \LTtwo{}. The hybrids' measured caches are $3$--$17\%$ larger than the analytic values in \autoref{tab:main-results}, because the kernel keeps the recurrent state in fp32 and we also count the short-convolution windows.

\paragraph{Speed}
The larger batches raise \LPT{}'s peak throughput above \LT{}'s by $7$--$91\%$, most at L with five recursions, where \LT{} fits only $21$ sequences. At equal batch, \LPT{} is $3$--$8\%$ slower than \LT{}, since each later recursion still reads the whole shared cache, and its own window too. As \citet{brandon2024crosslayerattention} note for sharing across layers, a shared cache saves memory, not the time to read it. Both looped models pay for depth in time: at batch $1$, a decoding step of \LT{} takes about $2.3$, $3.4$, and $5.7$ times as long as \Std{}'s at two, three, and five recursions, and \LPT{}'s takes $8$--$12\%$ longer than \LT{}'s. The hybrids read a full-context cache in only one layer of five, so they decode faster. At its maximum batch, \HLPT{} generates $2.1$--$2.2$ times as many tokens per second as \Std{} with two recursions, $1.4$--$1.5$ times with three, and $0.8$--$0.9$ times with five, and it is $2$--$3\%$ slower than \LTtwo{} at equal batch.

\begin{figure}[!ht]
    \centering
    \includegraphics[width=\linewidth]{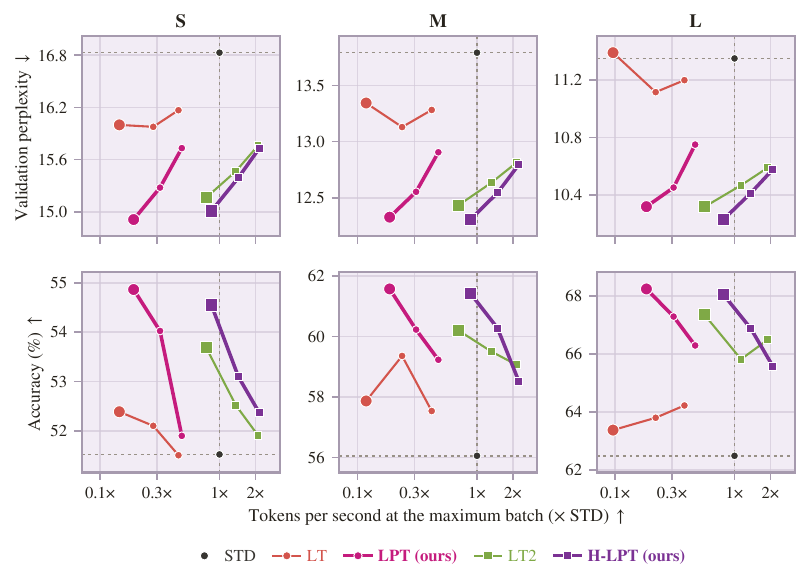}
    \caption{\textbf{\HLPT{} is better than \Std{} at similar or higher throughput.} Validation perplexity (top) and mean five-task accuracy (bottom) against generated tokens per second at each model's maximum batch, relative to same-size \Std{}, on one H100 at a $4{,}096$-token context. Curves follow two, three, and five recursions, with larger markers at five. Dashed lines mark \Std{}.}
    \label{fig:throughput-quality}
\end{figure}

\paragraph{Throughput and Quality}
\autoref{fig:throughput-quality} plots quality against throughput. At every size and recursion count, \LPT{} has higher throughput, lower perplexity, and higher accuracy than \LT{}, and \HLPT{} has higher throughput and lower perplexity than \LTtwo{}. Against \Std{}, \HLPT{} lowers perplexity by $0.94$--$1.44$ with three recursions while decoding $1.4$--$1.5$ times as many tokens per second, and by $1.12$--$1.82$ with five, at $81$--$88\%$ of \Std{}'s throughput.

\begin{table}[!ht]
    \centering
    \setlength{\abovecaptionskip}{0pt}\setlength{\belowcaptionskip}{6pt}
    \caption{\textbf{Decoding capacity and speed on one H100.} One table per size. bf16, $4{,}096$-token context. Batch: the largest batch that decodes without running out of the $80$ GB. Tok/s: generated tokens per second at that batch. Lat.: milliseconds per decoding step at batch $1$. Our rows are shaded.}
    \label{tab:wallclock}
    \sisetup{mode=text, group-separator={,}, group-minimum-digits=4}
    \setlength{\tabcolsep}{3.5pt}
    \sbox0{%
    \begin{tabular}{@{}lS[table-format=4.0]S[table-format=5.0]S[table-format=2.1]S[table-format=4.0]S[table-format=5.0]S[table-format=2.1]S[table-format=4.0]S[table-format=5.0]S[table-format=2.1]S[table-format=4.0]S[table-format=5.0]S[table-format=2.1]@{}}
        \toprule
        \textbf{S} & \multicolumn{3}{c}{$\numrec=1$} & \multicolumn{3}{c}{$\numrec=2$} & \multicolumn{3}{c}{$\numrec=3$} & \multicolumn{3}{c}{$\numrec=5$} \\
        \cmidrule(lr){2-4} \cmidrule(lr){5-7} \cmidrule(lr){8-10} \cmidrule(lr){11-13}
        Model & {Batch} & {Tok/s} & {Lat.} & {Batch} & {Tok/s} & {Lat.} & {Batch} & {Tok/s} & {Lat.} & {Batch} & {Tok/s} & {Lat.} \\
        \midrule
        \Std{} & 552 & 18570 & 2.1 &  &  &  &  &  &  &  &  &  \\
        \LT{} &  &  &  & 275 & 8414 & 4.8 & 184 & 5172 & 7.1 & 110 & 2688 & 11.8 \\
        \rowcolor{lptrow}
        \textbf{\LPT{}} &  &  &  & 540 & 8975 & 5.2 & 533 & 5894 & 7.9 & 514 & 3544 & 13.3 \\
        \LTtwo{} &  &  &  & 1513 & 39006 & 2.6 & 1007 & 25470 & 3.9 & 603 & 14348 & 6.4 \\
        \rowcolor{lptrow}
        \textbf{\HLPT{}} &  &  &  & 2746 & 40066 & 2.7 & 2531 & 26692 & 4.0 & 2190 & 15900 & 6.7 \\
        \bottomrule
    \end{tabular}%
    }\ifdim\wd0>\linewidth\resizebox{\linewidth}{!}{\usebox0}\else\usebox0\fi
    \par\vspace{7pt}
    \sbox0{%
    \begin{tabular}{@{}lS[table-format=4.0]S[table-format=5.0]S[table-format=2.1]S[table-format=4.0]S[table-format=5.0]S[table-format=2.1]S[table-format=4.0]S[table-format=5.0]S[table-format=2.1]S[table-format=4.0]S[table-format=5.0]S[table-format=2.1]@{}}
        \toprule
        \textbf{M} & \multicolumn{3}{c}{$\numrec=1$} & \multicolumn{3}{c}{$\numrec=2$} & \multicolumn{3}{c}{$\numrec=3$} & \multicolumn{3}{c}{$\numrec=5$} \\
        \cmidrule(lr){2-4} \cmidrule(lr){5-7} \cmidrule(lr){8-10} \cmidrule(lr){11-13}
        Model & {Batch} & {Tok/s} & {Lat.} & {Batch} & {Tok/s} & {Lat.} & {Batch} & {Tok/s} & {Lat.} & {Batch} & {Tok/s} & {Lat.} \\
        \midrule
        \Std{} & 206 & 6478 & 4.2 &  &  &  &  &  &  &  &  &  \\
        \LT{} &  &  &  & 103 & 2703 & 9.7 & 68 & 1527 & 14.5 & 41 & 763 & 24.1 \\
        \rowcolor{lptrow}
        \textbf{\LPT{}} &  &  &  & 202 & 3074 & 10.5 & 199 & 2000 & 16.0 & 192 & 1201 & 27.0 \\
        \LTtwo{} &  &  &  & 563 & 13804 & 5.3 & 377 & 8511 & 8.0 & 224 & 4502 & 13.2 \\
        \rowcolor{lptrow}
        \textbf{\HLPT{}} &  &  &  & 1029 & 14549 & 5.5 & 949 & 9622 & 8.2 & 821 & 5706 & 13.6 \\
        \bottomrule
    \end{tabular}%
    }\ifdim\wd0>\linewidth\resizebox{\linewidth}{!}{\usebox0}\else\usebox0\fi
    \par\vspace{7pt}
    \sbox0{%
    \begin{tabular}{@{}lS[table-format=4.0]S[table-format=5.0]S[table-format=2.1]S[table-format=4.0]S[table-format=5.0]S[table-format=2.1]S[table-format=4.0]S[table-format=5.0]S[table-format=2.1]S[table-format=4.0]S[table-format=5.0]S[table-format=2.1]@{}}
        \toprule
        \textbf{L} & \multicolumn{3}{c}{$\numrec=1$} & \multicolumn{3}{c}{$\numrec=2$} & \multicolumn{3}{c}{$\numrec=3$} & \multicolumn{3}{c}{$\numrec=5$} \\
        \cmidrule(lr){2-4} \cmidrule(lr){5-7} \cmidrule(lr){8-10} \cmidrule(lr){11-13}
        Model & {Batch} & {Tok/s} & {Lat.} & {Batch} & {Tok/s} & {Lat.} & {Batch} & {Tok/s} & {Lat.} & {Batch} & {Tok/s} & {Lat.} \\
        \midrule
        \Std{} & 108 & 3162 & 6.7 &  &  &  &  &  &  &  &  &  \\
        \LT{} &  &  &  & 54 & 1203 & 15.3 & 36 & 691 & 22.8 & 21 & 303 & 37.9 \\
        \rowcolor{lptrow}
        \textbf{\LPT{}} &  &  &  & 106 & 1480 & 16.4 & 104 & 976 & 25.0 & 101 & 578 & 42.3 \\
        \LTtwo{} &  &  &  & 258 & 5986 & 8.8 & 172 & 3585 & 13.1 & 103 & 1776 & 21.7 \\
        \rowcolor{lptrow}
        \textbf{\HLPT{}} &  &  &  & 475 & 6612 & 9.0 & 444 & 4333 & 13.6 & 390 & 2553 & 22.6 \\
        \bottomrule
    \end{tabular}%
    }\ifdim\wd0>\linewidth\resizebox{\linewidth}{!}{\usebox0}\else\usebox0\fi
\end{table}

\end{document}